\documentclass[a4paper,fleqn]{cas-sc}

\usepackage[authoryear]{natbib}
\usepackage{adjustbox}
\usepackage{multirow}
\usepackage{longtable}
\usepackage{geometry} 
\usepackage{tabularx}
\usepackage{booktabs}
\usepackage{subcaption}
\usepackage{booktabs}

\def\tsc#1{\csdef{#1}{\textsc{\lowercase{#1}}\xspace}}
\tsc{WGM}
\tsc{QE}
\tsc{EP}
\tsc{PMS}
\tsc{BEC}
\tsc{DE}

\begin{document}
\let\WriteBookmarks\relax
\def\floatpagepagefraction{1}
\def\textpagefraction{.001}
\shorttitle{Intelligent Condition Monitoring}


\title [mode = title]{Intelligent Condition Monitoring of Industrial Plants: An Overview of Methodologies and Uncertainty Management Strategies}

\author[1]{Maryam Ahang}[type=editor,
                        auid=000,bioid=1,]
\ead{maryamahang@uvic.ca}
\credit{Conceptualization, Formal analysis, Investigation, Data Curation, Visualization, Writing - Original Draft}

\author[1]{Todd Charter}[orcid=0000-0001-5982-255X]
\ead{toddch@uvic.ca}
\credit{Conceptualization, Writing - Original Draft, Writing - Review \& Editing}

\author[2]{Mostafa Abbasi}
\ead{abbasi@uvic.ca}
\credit{Visualization, Writing - Review}

\author[2]{Maziyar Khadivi}
\ead{mazy1996@uvic.ca}
\credit{Conceptualization}

\author[2]{Oluwaseyi Ogunfowora}
\ead{ogunfool@uvic.ca}
\credit{Conceptualization, Data Curation}

\author[1,2]{Homayoun Najjaran}
\cormark[1]
\ead{najjaran@uvic.ca}
\credit{Writing - Review \& Editing, Supervision}

\address[1]{Department of Electrical and Computer Engineering, University of Victoria, Victoria, BC, V8P 5C2, Canada}

\address[2]{Department of Mechanical Engineering, University of Victoria, Victoria, BC, V8P 5C2, Canada}

\cortext[cor1]{Corresponding author}

\begin{abstract}
Condition monitoring is essential for ensuring the safety, reliability, and efficiency of modern industrial systems. With the increasing complexity of industrial processes, artificial intelligence (AI) has emerged as a powerful tool for fault detection and diagnosis, attracting growing interest from both academia and industry. This paper provides a comprehensive overview of intelligent condition monitoring methods, with a particular emphasis on chemical plants and the widely used Tennessee Eastman Process (TEP) benchmark. State-of-the-art machine learning (ML) and deep learning (DL) algorithms are reviewed, highlighting their strengths, limitations, and applicability to industrial fault detection and diagnosis. Special attention is given to key challenges, including imbalanced and unlabeled data, and to strategies by which models can address these issues. Furthermore, comparative analyses of algorithm performance are presented to guide method selection in practical scenarios. This survey is intended to benefit both newcomers and experienced researchers by consolidating fundamental concepts, summarizing recent advances, and outlining open challenges and promising directions for intelligent condition monitoring in industrial plants.



\end{abstract}




\begin{keywords}
Condition monitoring \sep Fault detection and diagnosis \sep Machine learning \sep Deep learning \sep Industrial plants
\end{keywords}

\maketitle

\section{Introduction}

\begin{figure}[htbp]
	\centering
		\includegraphics[width=0.7\textwidth]{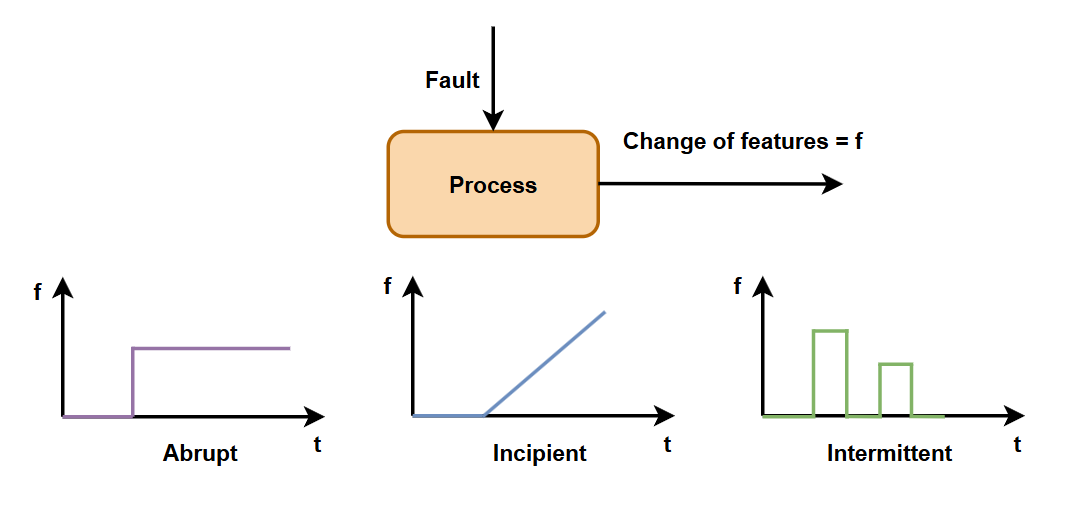}
	\caption{Time-dependency of faults \cite{isermann2005model}.}
	\label{FIG:0}
\end{figure}

Condition monitoring (CM) is fundamental to industrial systems, enabling early detection of faults and supporting efficient, safe, and reliable operations \cite{shojaeinasab2022intelligent}. Within this context, fault detection and diagnosis (FDD) reinforces process safety by enabling the timely isolation of system faults, helping to prevent escalation into incidents that could affect human safety or the environment. In industrial systems, CM can be defined as the process of analyzing measurable variables to identify significant changes that are signs of a developing fault. A fault is an unpermitted deviation of the characteristic property of a variable from standard behavior \cite{isermann2005model}. Faults can be categorized into three classes based on their time dependency: abrupt fault (stepwise), incipient fault (drift-like), and intermittent fault (usually occurs at irregular intervals), shown in Figure \ref{FIG:0}. If the fault is not detected and corrected on time, it may lead to a failure, which is a permanent interruption in the system's performance. The tasks of FDD have various definitions in different literature; however, in this paper, we define these tasks as fault detection, fault classification, fault isolation, and fault identification, based on \cite{isermann2005model} and \cite{izadi2006fault}. 


\emph{Fault detection} is the task of determining whether a fault has occurred in the system or not.
\emph{Fault classification} determines the type of faults and discriminates between different groups of faults.
\emph{Fault diagnosis} identifies the type of faults with as much specificity as possible, including problem size, location, and time of detection.
\emph{Fault isolation} determines the location of each fault, for example, which part of the system has become faulty.
\emph{Fault identification} determines the type, magnitude, and cause of the fault.
In general, FDD methods can be divided into three main categories: \emph{hardware redundancy, model-based fault detection,} and \emph{signal-based fault detection}.
Hardware redundancy is a method in which multiple sensors, hardware, and software are used to measure variables. With these measurements, final FDD decisions can be made based on a weighted vote among the collected information. This method is highly reliable and common in very sensitive systems; however, it requires additional space to place the redundant hardware and can also be expensive.
In model-based fault detection, faults are detected by comparing the available measurements of the system in conjunction with information from a mathematical model of the system. A residual amount is determined using the difference between the actual measurements and the estimates obtained from the model. Then, by applying a threshold to this residual quantity, a fault can be detected. Although this method is reliable, defining a precise model of a complex system is not an easy task, posing a challenge even to domain experts. On the other hand, signal-based methods rely solely on the analysis of data collected from sensors to detect deviations from normal behavior and are the most frequently employed FDD methods. This approach is useful for monitoring complex systems where traditional approaches may not be sufficient, and it has been widely applied to many different industrial systems and processes. As there are numerous sensors and data in any industrial system, deep learning algorithms can be utilized to handle such information. However, this approach is highly dependent on the quality and availability of the data. ML algorithms are widely used for condition monitoring \cite{surucu2023condition}. Moreover, combining the mentioned FDD methods, using hybrid approaches, and integrating prior knowledge into the data-driven and ML methods will enhance the accuracy and safety of the process.

\par Performance and evaluation metrics serve as valuable tools for comparing and evaluating fault detection and diagnosis capabilities of different methods. This section describes the most important and widely used metrics within the context of industrial CM to provide sufficient knowledge for comparing such studies throughout the remainder of this paper. Many of these studies include the classification terms True Positive (TP), False Positive (FP), True Negative (TN), and False Negative (FN). In this study, we treat the positive class as fault / abnormal and the negative class as normal:
\begin{itemize}
    \item \textbf{TP} (true positives): samples that are actually \emph{faulty} and predicted as faulty.
    \item \textbf{TN} (true negatives): samples that are actually \emph{normal} and predicted as normal.
    \item \textbf{FP} (false positives): samples that are actually \emph{normal} but predicted as faulty (false alarms).
    \item \textbf{FN} (false negatives): samples that are actually \emph{faulty} but predicted as normal (missed detections).
\end{itemize} It is crucial to achieve accurate detection of both normal and faulty samples. However, minimizing false negatives is considered a more important task, as it ensures that no fault condition is mistakenly considered normal. If there is any abnormality, this prioritization enables prompt corrective action. The most used performance metrics are shown in Table \ref{tab:performance_metrics}. 

\begin{table}[h!]
\centering
\caption{Common performance metrics for fault detection and diagnosis}
\label{tab:performance_metrics}
\renewcommand{\arraystretch}{1.3}
\resizebox{0.9\textwidth}{!}{
\setlength{\tabcolsep}{8pt}
\begin{tabular}{l c p{8cm}}
\toprule
\textbf{Metric} & \textbf{Formula} & \textbf{Description and Notes} \\ 
\midrule

Accuracy & 
$\displaystyle \frac{TP + TN}{TP + TN + FP + FN}$ & 
Measures the overall correctness of the model.Proportion of correctly classified samples out of all samples. It may be misleading for highly imbalanced datasets. \\

Precision & 
$\displaystyle \frac{TP}{TP + FP}$ & 
Indicates how many of the predicted faulty conditions were actually faulty. Higher precision means fewer false alarms. \\

Recall (Sensitivity) & 
$\displaystyle \frac{TP}{TP + FN}$ & 
Measures the model's ability to find all actual faulty conditions (true positive rate). Useful in safety-critical systems. \\

False Discovery Rate (FDR) & 
$\displaystyle \frac{FP}{TP + FP}$ & 
 The proportion of false alarms among all positive predictions. It's the complement of precision. A low FDR is desirable as it indicates a lower rate of erroneous fault detections. \\

False Alarm Rate (FAR) & 
$\displaystyle \frac{FP}{FP + TN}$ & 
The proportion of actual healthy conditions that were incorrectly identified as faulty. Lower values indicate better reliability (also known as false positive rate). \\

F1-score & 
$\displaystyle 2 \times \frac{\text{Precision} \times \text{Recall}}{\text{Precision} + \text{Recall}}$ & 
Harmonic mean of precision and recall. Recommended for imbalanced datasets. \\

\bottomrule
\end{tabular}
}
\end{table}

The focus of this paper is on studying different state-of-the-art deep learning algorithms that have been used for the condition monitoring of industrial plants with a focus on chemical processing plants. A comparison between different approaches is established through the analysis of their application on the Tennessee Eastman Process (TEP) simulator benchmark. This paper covers different deep learning methods for condition monitoring of industrial plants, compares and summarizes these methods, discusses the common problems of data-driven condition monitoring, and introduces solutions to address them.


Condition monitoring is an inevitable task in industrial chemical systems, and as such, numerous studies have been conducted on this topic. Yin et al. \cite{yin2012comparison} compared basic data-driven fault diagnosis and process monitoring methods such as Principal component analysis (PCA), DPCA, two variants of Partial least squares (PLS), Fisher discriminant analysis (FDA), and Independent component analysis (ICA) on the TEP benchmark. Their study compares both the performance and the computational complexity of the examined methods. Another study by Md Nor et al. \cite{md2020review} provides a comprehensive review of data-driven condition monitoring methods in chemical process systems. The research covers multivariate statistical analysis such as PCA, PLS, ICA, and FDA. In addition, the work also examines machine learning approaches such as artificial neural networks (ANNs), neuro-fuzzy (NF) methods, support vector machines (SVMs), Gaussian mixture models (GMMs), K-nearest neighbours algorithm (KNN), and Bayesian networks (BNs).

Limited studies have reviewed data-driven and machine-learning FDD methods for complex industrial processes \cite{yin2012comparison,md2020review, yin2014review, bi2022one}. However, to the best of our knowledge, there is no comprehensive study on deep learning and neural network-based CM approaches that fairly compares the methods based on a common benchmark case study. In this research, intelligent CM algorithms have been thoroughly analyzed and compared within the context of industrial chemical processes, covering implementation challenges such as the explainability of AI methods and data scarcity. Common challenges in data-driven condition monitoring of industrial systems, such as unlabeled, unseen, and imbalanced data, and how to tackle them, are introduced. Solutions for dealing with different types of data limitations and the explainability of ML methods are also explored.  Additionally, this work considers the current state and future directions of research within the broader scope of industrial CM. This research offers value to both new researchers and experts in the field. It introduces fundamental concepts for those new to intelligent condition monitoring while providing a comprehensive comparison of recent advancements and a discussion of challenges and limitations to guide future research.

\section{Research Methodology}

\par This study focuses on signal-based approaches for condition monitoring of industrial plants. , where the availability of rich sensor data in real-world applications motivates the use of machine learning (ML) and deep learning (DL) techniques.  In particular, neural network–based methods have attracted increasing attention due to their ability to process high-dimensional data and capture complex nonlinear relationships.  To provide a consistent basis for comparison, the Tennessee Eastman Process (TEP), a widely used benchmark in process systems engineering, is selected as the primary case study. Performance of different condition monitoring methods is reviewed in the context of this benchmark, highlighting both their strengths and limitations. 
A systematic literature search was conducted using a three-level keyword assembly designed to capture research on intelligent condition monitoring, AI techniques, and benchmark applications. The search query using All Fields operators was formulated as follows:
\\

("Condition Monitoring" OR "Fault Detection" OR "Fault Diagnosis" OR "Prediction", "Fault Prognosis" OR "Anomaly Detection") AND ("Machine Learning" OR "Deep Learning" OR "Neural Networks") AND ("Tennessee Eastman").
\\

\begin{figure}[h!]
\centering
\begin{subfigure}[t]{0.48\textwidth}
    \centering
    \includegraphics[height=5cm]{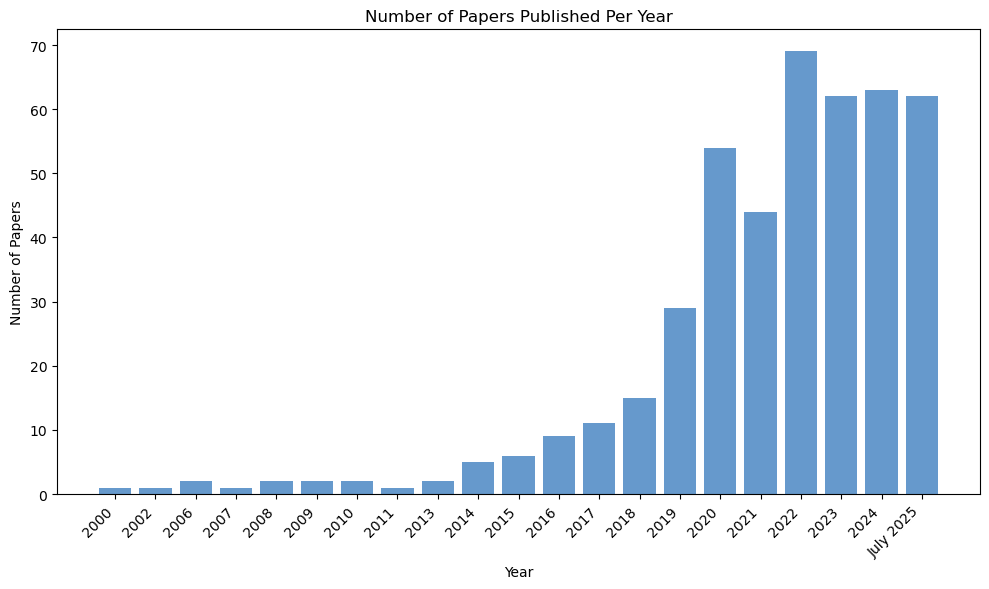}
    \caption{Number of publications over the years.}
    \label{FIG:year}
\end{subfigure}
\hfill
\begin{subfigure}[t]{0.48\textwidth}
    \centering
    \includegraphics[height=5cm]{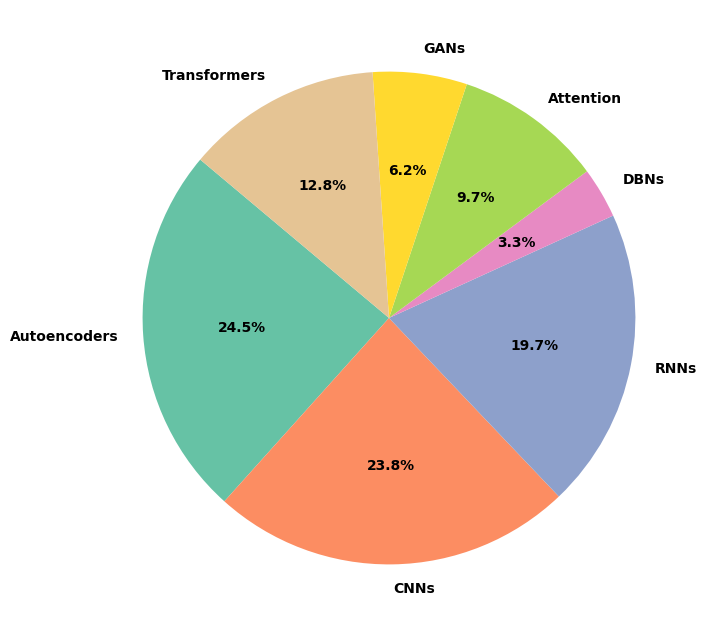}
    \caption{Visual taxonomy of methods.}
    \label{FIG:nns}
\end{subfigure}
\caption{Review metrics.}
\label{fig:metrics_and_taxonomy}
\end{figure}

Using this formulation, more than 450 papers were retrieved from the Web of Science (WoS) database. The inclusion criteria required papers to be written in English,  focused on ML/DL-based signal-level condition monitoring of industrial plants. Exclusion criteria included papers that focused solely on traditional non-neural ML methods (e.g., PCA, SVM) without DL integration, or lacked sufficient methodological detail for meaningful comparison.
The temporal distribution of the selected papers is shown in Figure \ref{FIG:year}. While early applications of ML to condition monitoring emerged in the early 2000s, interest in advanced DL methods accelerated after 2015, largely due to algorithmic breakthroughs, improved computational resources, and the availability of large-scale industrial datasets.
Within the scope of ML and DL, the most frequently employed algorithms include Autoencoders (AEs), Convolutional Neural Networks (CNNs), Recurrent Neural Networks (RNNs), Deep Belief Networks (DBNs), attention mechanisms, and Generative Adversarial Networks (GANs). Their distribution across the exdtracted papers is illustrated in Figure \ref{FIG:nns}. These categories form the backbone of the comparative analysis presented in this review.




\section{The Evolution of Intelligent Condition Monitoring }

This section explores some of the most important algorithms that have been successfully used for the condition monitoring of industrial plants, with a particular focus on data-driven methods. These approaches leverage machine learning techniques to detect and diagnose faults directly from process signals, often without relying on detailed physical models. While the emphasis is on recent advances, we begin with a brief overview of foundational methodologies, including probabilistic reasoning and hybrid diagnostic systems, which have shaped the development of modern intelligent monitoring frameworks.

\subsection{Early Methodologies and Foundations}
Early work in condition monitoring laid the foundation for modern data-driven fault detection strategies through the use of interpretable, statistically grounded models. Among the most influential were multivariate statistical process monitoring techniques such as PCA and PLS, which enabled unsupervised and supervised monitoring of high-dimensional process data \cite{kresta1991multivariate, wilson2000pls}. Extensions like Dynamic PCA \cite{ku1995disturbance} and Multiway PCA \cite{nomikos1994monitoring} introduced temporal structure, improving detection sensitivity for dynamic and batch processes.

As limitations of linear models became evident, researchers explored nonlinear extensions and hybrid methods. For example, Nonlinear Dynamic PCA combined neural networks with PCA to model nonlinear dynamics \cite{lin2000nonlinear}, while Canonical Variate Analysis (CVA) offered improved fault sensitivity in time-dependent settings \cite{russell2000fault}. Hybrid approaches, such as the combination of signed directed graphs with dynamic PLS, allowed for structured fault isolation and improved diagnosis of multiple simultaneous faults \cite{lee2004multiple}. These methods emphasized the value of integrating statistical learning with process-specific knowledge.

Probabilistic and knowledge-informed techniques also emerged during this period. Early applications of Bayesian reasoning, expert systems, and qualitative models provided interpretable and explainable fault analysis \cite{venkatasubramanian2003review}. Notably, Kulkarni et al. demonstrated that embedding domain insights into support vector machines improved classification accuracy, suggesting a transition toward learning approaches informed by prior knowledge \cite{kulkarni2005knowledge}.

The ideas introduced by these early approaches continue to influence the development of more advanced monitoring frameworks. In the following sections, we examine how recent advances build upon and extend these early contributions to address increasingly complex monitoring challenges.

\subsection{Prominent Process Monitoring Benchmarks}

\par Several simulators and datasets serve as benchmarks for process condition monitoring, including the Tennessee Eastman Process (TEP), PenSim, RAYMOND, DAMADICS, IndPenSim, FCC fractionator, Cranfield Multiphase Flow Facility \cite{melo2022open}. In this research, TEP is utilized as the primary benchmark for comparison due to its widespread acceptance and use in numerous studies. Some of the other case studies are examined in the section titled Industrial Applicability Beyond TEP.

\begin{figure}
	\centering
		\includegraphics[scale=.55]{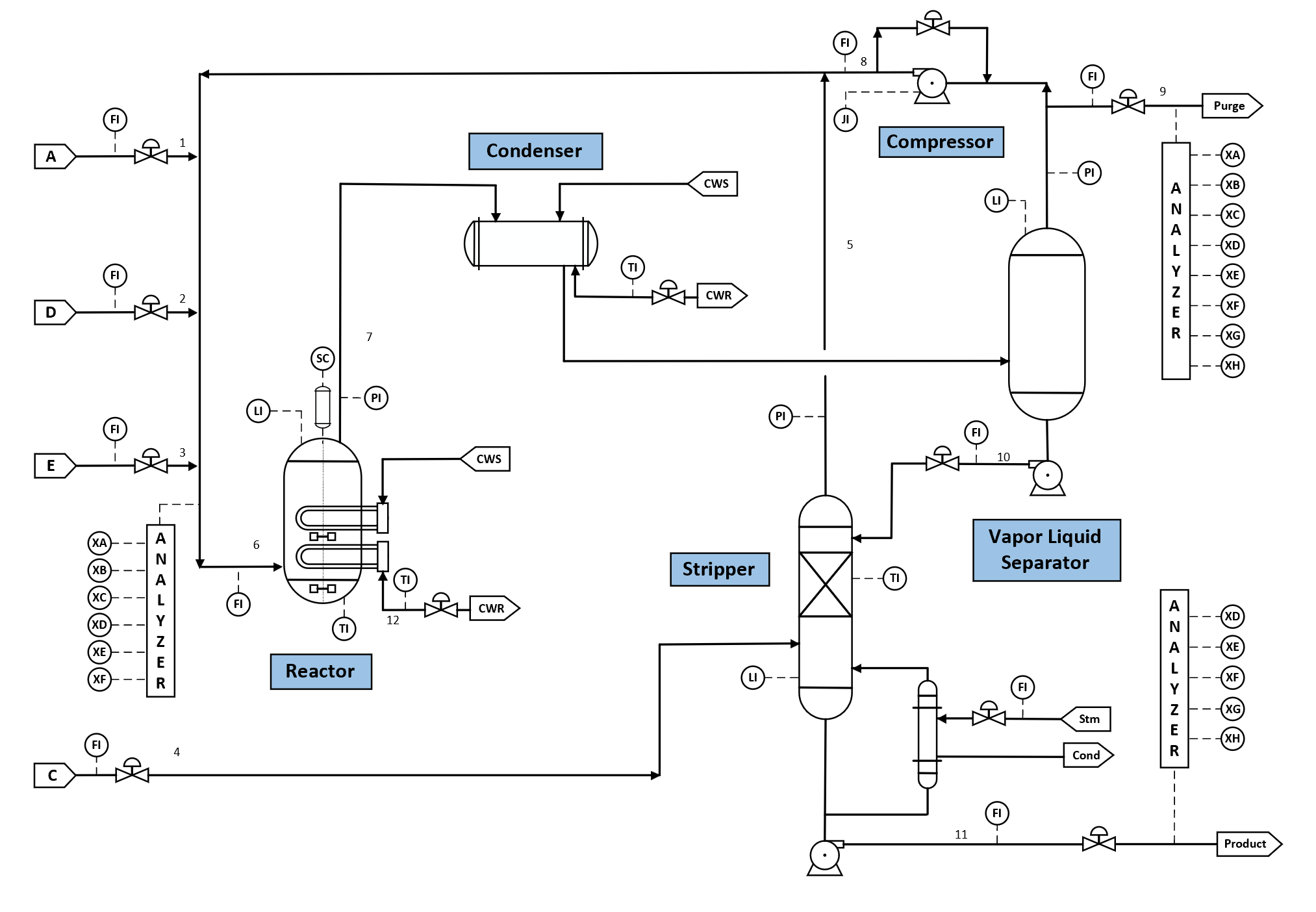}
	\caption{Tennessee Eastman Process flow diagram adapted from \cite{downs1993plant}} 
	\label{FIG:1}
\end{figure}

TEP is a realistic simulation of an industrial chemical plant, which is the most used benchmark for process control and monitoring studies \cite{melo2022open,yin2012comparison}. The process is described in \cite{downs1993plant}, and both the Fortran and Matlab simulation codes, along with numerous generated datasets, can be found online. The process consists of five main units: the reactor, the product condenser, a vapour-liquid separator, a recycle compressor, and a product stripper. The flow diagram of the process is shown in Figure \ref{FIG:1}. The process produces two products from four reactants, along with an inert and a byproduct. This results in a total of eight components denoted as A, B, C, D, E, F, G, and H. The reactions are shown as follows:
\begin{flalign}\label{TEP eqs}
\begin{split}
 A_{(g)}+C_{(g)}+D_{(g)} \rightarrow G_{(l)}
 \\
 A_{(g)}+C_{(g)}+E_{(g)} \rightarrow H_{(l)}
\\
 A_{(g)}+E_{(g)} \rightarrow F_{(l)}
\\
 3D_{(g)} \rightarrow 2F_{(l)}
\end{split}
\end{flalign}

The original simulation consists of 41 process variables and 12 manipulated variables, with sampling intervals of 3, 6, and 15 minutes. A total of 21 process faults, including step, random variation, slow drift, and sticking faults,  are defined for this simulation \cite{yin2012comparison}. Refer to Table \ref{tab:TEPFaults} for details.  Some faults, like 3, 9, and 15, are harder to detect.

\begin{center}
\begin{table}[]
\caption{Descriptions of process faults in TE process.}\label{tab:TEPFaults}

\begin{tabular}{ccc}
\hline
Fault Number & Process variable                                                                      & Type              \\ \hline
IDV(1)       & \begin{tabular}[c]{@{}c@{}}A/C feed ratio, B composition\\ constant\end{tabular}      & Step              \\
IDV(2)       & \begin{tabular}[c]{@{}c@{}}B composition, A/C ratio\\ constant\end{tabular}          & Step              \\
IDV(3)       & D feed temperature                                                                    & Step              \\
IDV(4)       & \begin{tabular}[c]{@{}c@{}}Reactor cooling water inlet\\ temperature\end{tabular}     & Step              \\
IDV(5)       & \begin{tabular}[c]{@{}c@{}}Condenser cooling water inlet\\ temperature\end{tabular}   & Step              \\
IDV(6)       & A feed loss                                                                           & Step              \\
IDV(7)       & \begin{tabular}[c]{@{}c@{}}C header pressure loss-reduced\\ availability\end{tabular} & Step              \\
IDV(8)       & A, B, and C feed composition                                                          & Random variation  \\
IDV(9)       & D feed temperature                                                                    & Random variation  \\
IDV(10)      & C feed temperature                                                                    & Random variation  \\
IDV(11)      & \begin{tabular}[c]{@{}c@{}}Reactor cooling water inlet\\ temperature\end{tabular}     & Random variation  \\
IDV(12)      & \begin{tabular}[c]{@{}c@{}}Condenser cooling water inlet\\ temperature\end{tabular}   & Random variation  \\
IDV(13)      & Reaction kinetics                                                                     & Slow drift        \\
IDV(14)      & Reactor cooling water valve                                                           & Sticking          \\
IDV(15)      & Condenser cooling water valve                                                         & Sticking          \\
IDV(16 -20)      & Unknown                                                                               & Unknown           \\
IDV(21)      & \begin{tabular}[c]{@{}c@{}}The valve fixed at steady state\\ position\end{tabular}    & Constant position \\ \hline
\end{tabular}
\end{table}
\end{center}

\subsection{Recurrent Neural Networks}

Recurrent neural networks (RNNs) are a class of deep learning models specifically designed to capture temporal dependencies in sequential data, making them well-suited for condition monitoring in dynamic industrial processes. Unlike feedforward networks, RNNs contain feedback connections that allow information to persist across time steps. However, standard RNNs often suffer from issues such as vanishing or exploding gradients during training. To address these limitations, Long Short-Term Memory (LSTM) networks were introduced with gated memory cells that enable learning over longer time horizons \cite{hochreiter1997long}. Gated Recurrent Units (GRUs) offer a more lightweight alternative to LSTMs by simplifying the gating mechanisms, often achieving comparable performance with reduced computational overhead. 




\par One of the first studies that used LSTM for chemical process fault diagnosis is \cite{zhao2018sequential}, in this case, no specific feature extraction is needed. Applying batch normalization makes the convergence of the network much faster.  To improve the diagnosis using LSTM networks, Han et al. \cite{han2020optimized} used an iterative method to optimize the number of hidden layer nodes.
Kang et al. \cite{kang2020visualization} used RNNs for fault classification. It is shown that RNN can create an even spatial distribution of fault types and classify them accurately. For TEP data, a 4-layer RNN could reach the average classification accuracy of 92.5\% for all faults; however, it can not detect fault 15. To overcome this problem, an LSTM network is used, and the average accuracy increased to  95\%. LSTM enhances RNNs by incorporating memory cells and gating mechanisms to better capture long-term dependencies.
To extract long-term dependencies in raw data and detect abnormal values, a Stacked Long Short-Term Memory network with three layers of LSTM is introduced in \cite{zhang2020novel}. As the historical information is stored in the memory, the monitoring system can operate in real time. The average FDR for all TEP faults is 99.49\%. Chadha et al. \cite{chadha2020bidirectional} used bidirectional recurrent neural networks for fault diagnosis, enabling the addition of more information about adjacent points into a training sequence than their unidirectional counterparts. They introduced a novel data preprocessing and restructuring for efficient network training. The average FDR for 21 faults of TEP is 88.14\%. To overcome nonlinearity and the high degree of spatio-temporal correlations, and incomplete characterization of the uncertainty in process noise and dynamics, a novel Bayesian recurrent neural network (BRNN) based FDI method is proposed in \cite{sun2020fault}, where we can have a probabilistic model of input-output mapping. This method can be used for fault propagation path analysis as well. 
To improve the feature extraction of unidirectional RNNs that only consider positive direction,  bidirectional RNN (BiRNN) models are used by \cite{zhang2019bidirectional}. Simple RNN unit (SRU), LSTM, and GRU are used in making bidirectional networks with the average FDR of 91.0\%, 92.3\%, and 92.7\%, respectively.

Mirzaei et al. \cite{mirzaei2022comparative} compared LSTM and GRU networks for FDD of chemical processes by using t-SNE visualization. GRU performed better than LSMT, especially for fault 15, which is very hard to detect. The average accuracies of LSTM and GRU for all faults are 95\% and 95.9\%, respectively.

Yuan et al. \cite{yuan2019intelligent} used GRU to extract the information from correlated sequential data; a moving horizon is used at the input of the GRU to divide the raw data into sequence units. To overcome the covariate shift and improve the network's learning efficiency, batch normalization is used. BN-based GRU is faster and more accurate. GRU and BN-GRU accuracy is 83.42\% and 87.36\%, respectively. An end-to-end model fusion feature learning method (MCNN-DBiGRU) is introduced in \cite{zhu_design_2025}, a feature-aligned multi-scale extraction model (MCNN) followed by a deep bidirectional mechanism that extracts time series features in the data, allowing the recurrent neural network to process input features both from past to future and from future to past. The experiments on all the fault samples of TEP achieve the precision of 94.11\%. 

A summary of recurrent models is shown in Table \ref{tab:RNN}. Recurrent architectures demonstrate strong capability in modeling the temporal dependencies inherent in process signals, making them well-suited for dynamic fault detection tasks. LSTM and GRU models consistently outperform vanilla RNNs due to their gating mechanisms, which mitigate vanishing gradients and enable longer memory retention. GRUs show comparable or slightly better performance in some instances, while offering reduced computational complexity, which is advantageous for real-time applications. However, recurrent models still face limitations in detecting rare faults (e.g., TEP fault 15) unless enhanced with bidirectional processing, attention mechanisms, or combined with convolutional feature extractors.

\begin{table}[h!]
\centering
\caption{Summary of Recurrent Models. See Table~\ref{tab:TEPFaults} for fault types.}
\label{tab:RNN}
\renewcommand{\arraystretch}{1.3}
\resizebox{0.9\textwidth}{!}{
\setlength{\tabcolsep}{5pt}
\begin{tabular}{p{0.16\textwidth}p{0.12\textwidth}p{0.4\textwidth}p{0.14\textwidth}p{0.08\textwidth}}
\toprule
\textbf{Model} & \textbf{Reference} & \textbf{Specification} & \textbf{Faults Information} & \textbf{Average Accuracy} \\
\hline

LSTM                & \cite{yuan2019multiscale}      & Discrete wavelet transform and CNN for feature extraction and softmax classifier                                                         & 1, 2, 4, 8, 10, 12, 13, 14, 17, and 18 & 98.73\%                 \\[0.6cm]
BiRNN               & \cite{zhang2019bidirectional}  & Bidirectional recurrent neural networks using gated recurrent unit (GRU)                                                                 & All                                    & 92.70\%                 \\[0.6cm]
LSTM + CNN          & \cite{shao2019multichannel}    & First LSTM and then CNN are used for feature extraction. The detection of faults 3, 9, and 15 are improved.                              & All                                    & 92.06\%                 \\[0.6cm]
RNN                 & \cite{kang2020visualization}   & Using a 4 layer RNN, however, could not detect fault 15                                                                                   & All                                    & 92.50\%                 \\[0.6cm]
LSTM                & \cite{kang2020visualization}   & Better performance compared to RNN speciallly for fault 15                                                                               & All                                    & 95\%                    \\[0.6cm]
LSTM                & \cite{zhang2020novel}          & Three layers LSTM are stacked                                                                                                            & All                                    & 99.49\%                 \\[0.6cm]
BiRNN               & \cite{chadha2020bidirectional} & Bidirectional recurrent neural networks                                                                                                  & All                                    & 88.14\%                 \\[0.6cm]
GRU                 & \cite{mirzaei2022comparative}  & The detection of fault 15 is enhanced compared to LSTM                                                                                   & All                                    & 95.90\%                 \\[0.6cm]
 
\bottomrule
\end{tabular}
}
\end{table}

\subsection{Convolutional Neural Network (CNN)}

\par Convolutional Neural Networks (CNNs) were introduced in the late 1980s and initially used to detect patterns in images \cite{le1989handwritten}. These powerful feature extraction networks are widely used for fault detection.



One of the first applications of using CNN for TEP fault detection and diagnosis is introduced in \cite{wu2018deep}. In this approach, a deep convolutional neural network (DCNN) with dropout extracts features from time series data. The implemented method achieved an average FDR of 88.2\% for 20 TEP faults; however, it heavily relies on historical fault data samples and is less effective for highly complex chemical processes involving thousands of variables. 
Some modifications to the CNN network are done in \cite{chadha2019time}  to increase the generalization ability of the model, improving feature extraction abilities and reducing the overfitting, such as residual connections, inception module, and dilation convolution. With these improvements, the F1 score of hard detectable faults 3,9,13,15 is improved to 0.91. Frequency domain data contains valuable information that is usually neglected in the traditional FDD of chemical processes. In \cite{zhang2019amplitude}, the Fast Fourier Transform (FFT) is used for TEP feature extraction, and the output data is fed into a convolutional neural network, which is used as a binary classifier. The FDR of this model on TEP data, considering all faults except 3, 9, 15, 16, is 96.9\%. 
In CNN, all the features have the same importance in the classification process.  Multi-head attention CNN (MACNN) \cite{cui2020improved} are introduced to distinguish the importance of different features, where the extracted features from convolution layers are fed into the multi-head attention (MA) layers, finally a softmax classifier is used to classify the different faults, the performance is improved compared to feedforward neural network (FNN) and CNN models. Multiblock temporal convolutional network (MBTCN) for FDD is introduced in \cite{he2021multiblock} using the idea of "local extraction and global integration"; it considers the cross-correlation and the temporal-correlation in the multivariate processes' data. In this method, most related process variables are divided into the same sub-blocks where TCN extracts the features. It achieved 96.2\% accuracy in TEP, considering all fault classes but 3,9,15. In \cite{yu2021multichannel}, a multichannel one-dimensional convolutional neural network (MC1-DCNN) was employed, using extracted features from the Wavelet transform.  The wavelet transform is a powerful tool for extracting time and frequency domain features. The average recognition accuracy of the implemented method for all 21 faults of TEP is  94.08\%. 
To handle multi-time scale data, Gao et al. \cite{gao2020process} introduced a feature extraction method based on CNN. The similarity measure is used to eliminate fault-irrelevant variables. The fixed multiple sampling (FMS) method is introduced for converting a feature map with an arbitrary shape into a fixed-length vector to handle different time scales of data. The average FDR for all 21 classes using the multi-time scale model is 98.53\%, while the single-time scale model is 94.65\%. A multi-model monitoring method is introduced in \cite{li2020wavelet}, in which first a CNN model detects some faults quickly and with minimum computational power, then a wavelet-assisted secondary CNN model accurately detects the remaining faults and also filters the noise and redundant information. The FDR of the model for 20 faults of TEP is 93\%. A normalized CNN is used in \cite{wang2021fault} to overcome the problem of gradient explosion and disappearance; moreover, the convergence speed of the network is improved. This network consists of the depth-normalized convolution network and improved second-order pooling for feature fusion. The average fault detection accuracy for all faults is 92.75\%. It should be noted that this method needs a large number of calculations. To deal with high-dimensional nonlinearity, the combination of matrix diagram and multi-scale convolutional neural network (MsCNN) is used in \cite{song2022multi}. Time series data is converted into matrix diagrams (2-D images). The average fault detection accuracy for all faults is 88.54\%. It should be noted that running this model is very time-consuming. Sparse one-dimensional convolutional neural network (S1-DCNN) \cite{yu2022sparse}, is introduced to extract features from complex signals, the method uses a sparsity regularization layer at the output of the convolution network to remove any redundant representations and extract fault-discriminative features. In \cite{qin2022adaptive}, an adaptive multiscale convolutional neural network (AMCNN), two attention modules, adaptive channel attention and adaptive spatial attention, are incorporated to select the most useful features and suppress redundant information.  Another novelty in this paper is the proposed optimization method using a weighted loss function consisting of classification loss, regularization loss, and triplet loss to improve robustness and generalization. The average fault detection rate for 20 faults of TEP is 93.02\%.  
Dynamic convolutional neural networks can extract temporal correlation of process data and are useful for feature selection and feature sequence arrangement, as in \cite{deng2021integrating}. First, the genetic algorithm (GA) is used for feature selection, and then CNN is used for fault diagnosis. The FDR for 20 faults of TEP is 89.72\%.
A 1D-CNN network is used in \cite{niu2022novel}, exploiting the correlation in the data; the average FDD accuracy for all TEP faults is 93\%. Multi-block adaptive convolution kernel neural network (MBCKN) \cite{guo2022multiblock} is a feature extraction method for large-scale processes, in which the process is decomposed into several blocks, and the attention mechanism is used for convolutional kernel selection for each block. Finally, an adaptive fusion method fuses the features of different blocks. This model can detect 21 faults of the TEP with an average accuracy of 94.15\%. To enhance feature extraction capabilities and prevent model degradation, a novel multi-scale residual jagged dilated convolution neural network (MRJDCNN) model is proposed in \cite{chen_reinforced_2024}. This model incorporates the principle of residual learning into an improved jagged dilated convolution and is then carefully combined with an LSTM network. This model could detect all the TEP faults with an average accuracy of 96.59\%. It is also validated on an industrial coking furnace.
A global–local attention convolutional neural network (GLACNN) is introduced in \cite{yang2025fault} for fault diagnosis in complex multi-unit nonlinear industrial processes, where traditional models struggle to balance intra-unit and inter-unit feature extraction. GLACNN enhances diagnostic performance by combining local features from individual sub-units with global features capturing system-wide interactions, using specially designed one-dimensional convolutional layers. Channel attention modules are employed to assign significance to features based on their relevance, improving interpretability and diagnostic accuracy. Validated on the TEP and penicillin fermentation processes, GLACNN demonstrates superior accuracy compared to baseline models, achieving an average FDR of 99.6\% for the first five fault types. However, its performance degrades when encountering unseen fault types common in dynamic industrial settings.

\subsubsection{LSTM and CNN}
Yuan et al. \cite{yuan2019multiscale} proposed a framework for feature extraction and fault diagnosis, where the discrete wavelet transform and CNN are used for feature extraction, then the LSTM network reduces the useless information, and finally classification is done by using a softmax classifier. The model is validated on a few faults (Fault 1, 2, 4, 8, 10, 12, 13, 14, 17, and 18) of the TEP, and the average F1-score is 0.9873.
In \cite{shao2019multichannel}, first, the LSTM network is used for feature extraction, then convolutional kernels extract characteristics of the data. This combination results in a high average fault detection accuracy of 92.06\% for the TEP data set. This model can detect faults 3, 9, and 15, which are very difficult to diagnose with a very high accuracy. Bao et al. \cite{bao2022chemical} used a deep convolutional neural network (DCNN) to extract spatial features from raw data and then fuse features into the bidirectional recurrent neural network (BRNN). No additional data preprocessing is needed in this case. The average accuracy for 20 faults of TEP is 91.7\%. The combination of the sliding window on integrated 2D image-like data of feature information and time delay information, and the CNN-LSTM model is used in \cite{huang2022novel} for process fault detection. This method is robust against noise and converges faster compared to separate CNN and LSTM networks. To ensure that no feature information is lost, Ren et al. \cite{ren_network_2024} used a CNN to extract multi-level fault features, combine them into a unified feature space, and form hyper features through feature fusion. These hyper features are then fed into a stacked LSTM network for additional feature extraction. This approach effectively captures time series information and addresses the issue of gradient disappearance during long-term training, achieving an accuracy of 95.19\% for all the faults. Zhao et al. \cite{zhao_deep_2024} used an adaptive multiscale CNN and enhanced highway LSTM (ACEL) to tackle the problem of diverse time scales and data features in industrial settings. This method starts by automatically extracting multiscale features under varying operating conditions through a parallel convolutional module. It then applies efficient channel attention to emphasize more important channels while reducing the influence of less relevant ones. A bidirectional LSTM is used to create fine-grained hybrid features, combining contextual and local information learned by the CNN. Additionally, an enhanced highway configuration is employed to strengthen global temporal dependencies. The FDR and F1 are more than 98\%. 

A summary of CNN-based methods is shown in Table \ref{tab:CNN}. Convolutional architectures are powerful in spatial feature extraction from multivariate process data and often outperform simpler feedforward networks. Variants such as MBTCN and MC1-DCNN demonstrate that domain-specific preprocessing (e.g., wavelet transforms, multiblock decomposition) significantly boosts detection rates for complex and multi-timescale processes. Attention-based CNNs (e.g., AMCNN) further improve fault-specific sensitivity by assigning higher weights to critical channels, addressing the equal importance limitation of plain CNNs. The main drawback is computational load, especially for combining CNN with multi-branch attention and multi-resolution networks which can be a barrier to online deployment in high-frequency monitoring scenarios.

\begin{table}[h!]
\centering
\caption{Summary of CNN Models. See Table~\ref{tab:TEPFaults} for fault types.}
\label{tab:CNN}
\renewcommand{\arraystretch}{1.3}
\resizebox{0.9\textwidth}{!}{
\setlength{\tabcolsep}{5pt}
\begin{tabular}{p{0.16\textwidth}p{0.12\textwidth}p{0.4\textwidth}p{0.14\textwidth}p{0.08\textwidth}}
\toprule
\textbf{Model} & \textbf{Reference} & \textbf{Specification} & \textbf{Faults Information} & \textbf{Average Accuracy} \\
\hline

DCNN                 & \cite{wu2018deep}              & CNN with dropout                                                                                                                         & 20 faults                              & 88.20\%                 \\[0.6cm]
CNN                 & \cite{zhang2019amplitude}      & FFT for feature extraction                                                                                                               & All faults except 3,9,15,16            & 96.90\%                 \\[0.6cm]
CNN                 & \cite{gao2020process}          & Handling multi time scale data using fixed multiple sampling method                                                                      & All                                    & 98.53\%                 \\[0.6cm]
MBTCN                 & \cite{he2021multiblock}        & Multiblock temporal convolutional network (MBTCN), dividing process variables into subblocks following the idea of local extraction and global integration                                            & All faults except 3,9,15               & 96.20\%                 \\[0.6cm]
MC1-DCNN                 & \cite{yu2021multichannel}      & 1D CNN and Wavelet transform for feature extraction                                                                                      & All                                    & 94.08\%                 \\[0.6cm]
Normalized CNN                & \cite{wang2021fault}           &  Improve the stability and speed of the training                                                                        & All                                    & 92.75\%                 \\[0.6cm]
MBCKN                 & \cite{guo2022multiblock}       & Decompose process into several blocks for large-scale process using attention mechanism                          & All                                    & 94.15\%                 \\[0.6cm]
CNN + LSTM          & \cite{ren_network_2024}        & CNN is used to extract multi-level fault features, followed by stacked LSTM solving the gradient disappearance problem.                    & All                                    & 95.19\%                 \\[0.6cm]
ACEL       & \cite{zhao_deep_2024}          & Extracts multiscale features for various operating conditions combining  CNN and LSTM                                                                          & All faults except 3,9,15,21            & 98\%                    \\[0.6cm]
MRJDCNN-LSTM          & \cite{chen_reinforced_2024}    & Improved feature extraction and prevent model degradation                                                                                & All faults                             & 96.59\%                    \\
 GLACNN  &\cite{yang2025fault} & Balancing intra-unit and inter-unit feature extraction, assigning significance to features, using channel attention modules, not ideal for unseen faults          &     Faults 1-5    & 99.6\%\\
\bottomrule
\end{tabular}
}
\end{table}

\subsection{Auto-encoders}  
\par Autoencoders are models first introduced in the late 1980s. They consist of two networks, the encoder and the decoder, where the encoder maps the input data to a hidden layer and the decoder tries to reconstruct the input data. In general, the process is done unsupervised \cite{schmidhuber2015deep}. They are powerful tools for feature extraction and dimension reduction. The output of the encoder is the feature representation and can be used for further analysis, such as the input of a classifier. 



\par Autoencoders and their variations \cite{ahang2024condition} have been widely used in condition monitoring of industrial plants where the available data is massive, specifically for chemical processes. One of the first attempts of using autoencoders for fault diagnosis of the TEP is done in \cite{lv2017weighted}, a stacked sparse autoencoder (SSAE) is implemented for extracting and learning high-level features and underlying fault patterns which is very useful for detecting incipient faults, time correlations among the samples is also considered, unlike other methods that have poor performance in diagnosing incipient faults, this algorithm could classify faults 3,9,15,21 of the TE which are very hard to detect with the accuracy of 95.75\%, 99.25\%, 98.75\%, 99.5\% respectively. Another powerful tool for fault detection, especially incipient faults in industrial plants, is the Deep LSTM Supervised Autoencoder Neural Network (Deep LSTM-SAE NN) \cite{agarwal2022hierarchical}. The assumption that having different models for different faults increases the sensitivity of the model is made in the implementation of this method. The average accuracy for all the classes is 93.23\%.

A stacked denoising autoencoder (SDAE) alongside the k-nearest neighbour (KNN) rule is used for automated feature learning in \cite{zhang2018automated}; the SDAE automatically extracts crucial features from nonlinear process data. Neural component analysis (NCA) for fault detection is introduced in \cite{zhao2018neural}, which is a unified model including a nonlinear encoder and a linear decoder. The network is trained with orthogonal constraints such as those used in PCA to alleviate the overfitting problem in autoencoders. In \cite{cheng2019novel}, a variational recurrent autoencoder is used for process fault detection. This method monitors the process in probability space to handle the nonlinearity more efficiently. Recurrent neural networks are also used to consider temporal dependency between variables. The negative variational score (NVS) is introduced as the monitoring metric, which considers reconstruction error and similarity between prior and posterior probability distributions. The model is validated on the TEP and the average fault detection rate of all faults except 3,9,15 is 96.3\%. The combination of an autoencoder and an LSTM is a powerful tool for fault detection of rare events. In \cite{park2019fault}, an autoencoder is trained with offline normal data and used as an anomaly detector. The predicted faults will be classified by using an LSTM network. The average fault detection accuracy for the first 20 faults is 91.9\%, and by not considering 9 and 15 faults, which are very hard to detect, the average accuracy is 96.8\%. The time delay of the fault detection of each class is also studied in this work. In \cite{yin2019mutual}, a fault detection model based on Mutual Information and Dynamic Stacked Sparse Autoencoder (SSAE) is introduced. The variables with a strong autocorrelation and their lag time of maximum autocorrelation were chosen using time series input data. Then, dynamic augmented data is fed to the SSAE for feature extraction and learning latent information in the dynamic process data. Based on selected process variables, the average accuracy of fault detection of all 21 faults is 85.06\%. Liu et al. \cite{liu2022toward} introduced a denoising sparse autoencoder (DSAE) to address the noise interference and nonlinearity in process variables. This method can achieve an accuracy higher than 99\% for some faults.
 
There are usually multiple operating modes in a chemical process, and fault detection models that are built based on one mode may have high false alarm rates. For multimode process monitoring and handling unseen modes, a self-adaptive deep learning method based on local adaptive standardization and variational auto-encoder bidirectional long short-term memory (LAS-VB) is introduced in \cite{wu2020self}. In this research, unstable tendencies in the local moving window are detected. This method is not efficient for very slow-drift faults. Combining autoencoders and clustering algorithms for fault detection is a powerful tool, especially for dealing with unlabeled data. Zheng et al. \cite{zheng2020new} used a convolutional stacked autoencoder (SAE) network for feature extraction. Feature visualization is done by the t-distributed stochastic neighbour embedding (t-SNE) algorithm and DBSCAN clustering. The implemented method could detect Faults 1, 2, 4, 6, 7, 11, 13, 14, 17, 19, and 20 with an average accuracy of 93\%. Clustering is done unsupervised, and a knowledge-based cluster annotation is used for labeling. 

Convolutional autoencoders are powerful feature extraction tools; however, the pertaining phase is done unsupervised. To improve the extracted features and make sure they contain internal spatial information, a supervised convolutional autoencoder (SCAE) is introduced \cite{yu2022supervised}. The average accuracy of this method on the TE dataset is 96.19\%, and it could reach an accuracy of about 95\% in only two epochs. One-dimension residual convolutional auto-encoder (1DRCAE) is a model introduced in \cite{yu2022one}, for unsupervised feature extraction, where two residual learning blocks are embedded between the encoder and decoder to improve the speed and accuracy. The fault detection rate of the model on the TEP dataset on all faults except 3, 9, and 15 is 90.8\%. 
Stacked spare-denoising autoencoder (SSDAE)-Softmax\cite{liu2021toward}, is introduced for robust fault detection in complex industrial processes. The state transition algorithm (STA) is a global optimization method that is used for hyperparameter optimization. The fault detection accuracy of the model on the TEP reaches an average of 96.6\% for faults 1, 2, 4, 5, 7, 13, and 17.
Conventional autoencoder algorithms typically prioritize the reconstruction error, overlooking valuable information in the latent space for FDD. To address this problem, a one-dimensional convolutional adversarial autoencoder (1DAAE) is introduced in \cite{wang2023novel}, which uses the adversarial thought for clustering the latent variables to a prior distribution. Both reconstruction error and latent variable distribution are considered in the training of the model. The mean accuracy for all 21 faults of TEP is 86\%. To monitor the nonlinear process and capture the process dynamics, a novel dynamic-inner convolutional autoencoder (DiCAE) is introduced in \cite{zhang2022dynamic}, integrating a vector autoregressive model into a 1-dimensional convolutional autoencoder. The average FDR of 96.8\% is achieved on the TEP for all the classes except 3, 9, and 15. Jang et al. \cite{jang2021adversarial}, combined variational autoencoder and a generative adversarial network proposing an adversarial autoencoder (AAE),  to generate features that follow the designed prior distribution. These features improve the reliability and stability of FDD algorithms. The average FDR for all 21 faults of TEP is 78.55\%. This method decreases the fault detection delay and can detect faults faster compared to VAE. To reduce the irrelevant information with the faults during feature extraction, a neuron-grouped stacked autoencoder (CG-SAE) is introduced in \cite{pan2021classification} hierarchical fault-related feature representation. In this method, hidden neurons are divided into different groups representing the data categories to extract category-related features. This method achieved the average FDR of 96.02\% for all faults except 3, 9, and 15. Li et al. \cite{li2020fault} introduced a multimode feature fusion algorithm that selects the most relevant variables for each fault using the minimum redundancy-maximum relevance (mRMR) method, and then uses a stacked autoencoder for feature extraction. The average FDR for all faults except 3,9, and 15 is 88.04\%, which is improved compared to SAE with the FDR of 76.77\%. To reduce the noise effect in high-dimensional process signals, a one-dimensional convolutional auto-encoder (1D-CAE) is introduced in \cite{chen2020one}.. This model performs feature extraction in a hierarchical structure, outperforming SVM, DBN, and SDAE. Jin et al. \cite{jin2018wavelons} utilized neurons with a wavelet activation function (Wavelons) and exponential linear units (ELU) to construct a Wavelons-constructed autoencoder-based deep neural network (WA-DNN), achieving an average FDR of 95.86\% across all 21 faults.

Traditional fault detection techniques do not take quality indicators into account; they only look at process data. A mixing stacked auto-encoder, consisting of a nonlinear encoder and a linear decoder, is introduced in \cite{yan2019design} to extract features of process variables and quality indicators while keeping feature extraction and model construction separate. Prior knowledge enhances the predictive accuracy of quality indicators and makes models easier to interpret, leading to more dependable monitoring performance. Stacked denoising auto-encoders (SDAE), are used to extract nonlinear latent variables and features from noisy data, however, the dynamic relationships are neglected. To consider them, a recursive stacked denoising auto-encoder (RSDAE) is proposed in \cite{zhu2020deep}. For extracting more effective features from time domain process signals, a convolutional long short-term memory autoencoder (CLSTM-AE) is introduced in \cite{yu2020convolutional}, where a residual block is used to select key information and improve the accuracy, and reduce gradient disappearance and gradient explosion in the training phase. FDR for all the faults of TEP is 91\%. Nonlocal and global preserving stacked autoencoder (NLGPSAE) \cite{yang2023nonlocal} can extract structure-related features from nonlinear process data by using a regularized objective function that can control the reconstructed space. The average FDR for all the faults of TEP, considering $T^2$ and SPE statistics, is 94.8\% and 91.8\%, respectively.


To consider geometrical information from process signals during the feature extraction of SAEs in fault detection, manifold regularized stacked autoencoders (MRSAE) are introduced in \cite{yu2020manifold} that consider the neighbourhood information. MRSAE does not need a large amount of labeled data. This model is superior in detecting certain types of faults like 5, 16, and 19, and the average FDR for 21 faults of TEP is 91.42\%. To achieve interpretable analysis along the fault detection task, a novel orthogonal self-attentive variational autoencoder (OSAVA) model is introduced in \cite{bi2021novel}. Orthogonal attention extracts the correlations between variables and the temporal dependency of different time steps. The self-attention mechanisms utilized in the VSAE consider information from all time steps to improve the performance. This model has a low detection delay, an average of 24.2 minutes, and the average FDR is  94.7\%. To achieve effective and interpretable features for condition monitoring, a slow feature analysis-aided autoencoder (SFA-AE) is introduced in \cite{li2021toward}, which extracts deep slow variation patterns from the extracted features of AE. These slow patterns can show abnormal states. The approach uses a neural network consisting of a process variable selection (PVS) block, followed by an encoder-decoder structure and a deep slow feature analysis (DSFA) section. The PVS block utilizes 1D convolutional layers to generate an attention map that represents feature importance for selecting crucial process variables. Next, the encoder and decoder sections use convolutional LSTM and deconvolutional LSTM layers, respectively, to learn spatiotemporal relationships of the input variables. The average FDR for 21 faults of TEP is 94.87\%.
To handle nonlinearity, non-Gaussian features, and dynamics in process monitoring of complex systems, independent component analysis (ICA) and adversarial autoencoder
(AAE) are fused in \cite{chen_new_2025}. To improve detection reliability, fault thresholds are determined using tail distribution features, achieving the average FDR of 99.52\% for all the faults except 3, 9, and 15. Residual squeeze-and-excitation convolutional auto-encoder (RSECAE) is introduced in \cite{yu2024residual} for effective feature extraction. This method proposes a hybrid structure combining convolution calculation, an auto-encoder, and multiple residual squeeze-and-excitation networks (RSENets), and utilizing maximum mean discrepancy (MMD) to reduce the distribution difference between the process signals and the learned features. This method is evaluated using TEP and a three-phase process (real industrial process), showing the ability of the unsupervised learning method in the absence of fault data.
A differential recurrent autoencoder (DRAE) is proposed in \cite{ji2025differential} to address the challenge of early fault detection in nonstationary industrial processes, where traditional methods struggle to capture transient, time-varying behaviors caused by stochastic disturbances and operational variability. By embedding differential operations into the latent layer of LSTM networks, DRAE enhances the forget gate's responsiveness to short-term nonstationary features while preserving long-term dependencies. The model leverages an unsupervised encoder-decoder structure, using reconstruction error as a fault indicator, effectively distinguishing early-stage faults from inherent nonstationary trends. Validation on synthetic data, a real-world chemical plant (industrial continuous catalytic reforming heat exchange unit), and TEP achieving the average FDR of 92.34\% for faults 1 to 14 except 3 and 9,  demonstrates that DRAE significantly improves fault detection rates and reduces detection delays compared to conventional methods.

Autoencoder-based methods are highly effective for unsupervised feature learning and dimensionality reduction, making them suitable when labeled fault data are scarce. Temporal integration (e.g., CLSTM-AE, OSAVA) boosts performance by capturing time-dependent variations in latent space, while supervised variations (e.g., SCAE) accelerate convergence and improve discriminative ability. Incorporating sparsity, manifold regularization, or adversarial learning helps enhance robustness against noise and improve separation of fault classes in latent space. However, autoencoders generally underperform in detecting extremely subtle or slowly developing faults unless paired with temporal models or variable selection mechanisms. Their success also heavily depends on the quality of normal training data to avoid false alarms. A summary of these methods is shown in \ref{tab:AE}.

\begin{table}[h!]
\centering
\caption{Summary of Autoencoder Models. See Table~\ref{tab:TEPFaults} for fault types.}
\label{tab:AE}
\renewcommand{\arraystretch}{1.3}
\resizebox{0.9\textwidth}{!}{
\setlength{\tabcolsep}{5pt}
\begin{tabular}{p{0.16\textwidth}p{0.12\textwidth}p{0.4\textwidth}p{0.14\textwidth}p{0.08\textwidth}}
\toprule
\textbf{Model} & \textbf{Reference} & \textbf{Specification} & \textbf{Faults Information} & \textbf{Average Accuracy} \\
\hline

WA-DNN         & \cite{jin2018wavelons}         & Using wavelet activation function in the autoencoder                                                                                     & All                                    & 95.86\%                 \\[0.6cm]
CLSTM-AE         & \cite{yu2020convolutional}     & Convolutional LSTM memory autoencoder (CLSTM-AE) and residual block for increasing the accuracy and making the training process more robust & All                                    & 91\%                    \\[0.6cm]
MRSAE         & \cite{yu2020manifold}          & Manifold regularized stacked autoencoders (MRSAE) considering neighbourhood information needs a few labeled data                         & All                                    & 91.42\%                 \\[0.6cm]
DiCAE         & \cite{zhang2022dynamic}        & integrating vector autoregressive model and 1-dimensional convolutional autoencoder for capturing the process dynamics                               & All faults except 3, 9, and 15          & 96.80\%                 \\[0.6cm]
CG-SAE         & \cite{pan2021classification}   & Neuron-grouped stacked autoencoder (CG-SAE) as a hierarchical feature extractor for different groups of faults                           & All faults except 3, 9, and 15          & 96.02\%                 \\[0.6cm]
OSAVA         & \cite{bi2021novel}             & Orthogonal self-attentive variational autoencoder (OSAVA) to increase the interpretability with a low detection delay                                              & All faults except 15, 16               & 94.70\%                 \\[0.6cm]
SFA-AE         & \cite{li2021toward}            & Slow feature analysis-aided autoencoder (SFA-AE) to deal with slow variations in the data                                                & All faults except 15, 17               & 94.87\%                 \\[0.6cm]
LSTM-SAE         & \cite{agarwal2022hierarchical} & Combining LSTM and Supervised Autoencoder, improving sensitivity by different models for different faults                                                                                                & All                                    & 93.23\%                 \\[0.6cm]
VAE + RNN        & \cite{cheng2019novel}          & Variational recurrent autoencoder combines RNN and VAE to handle nonlinearity                                                                                  & All faults except 3,9,15               & 96.30\%                 \\[0.6cm]
SCAE         & \cite{yu2022supervised}        & Supervised convolutional autoencoder (SCAE) for better feature extraction, high accuracy in a few epochs & All faults except 3,9,15,16            & 96.19\%                 \\[0.6cm]
1DAAE         & \cite{wang2023novel}           & Clustering the latent variables, considering both reconstruction error and latent space information                                                                                           & All                                    & 86\%                    \\[0.6cm]
NLGPSAE         & \cite{yang2023nonlocal}        & Nonlocal and global preserving stacked autoencoder (NLGPSAE) using a regularized objective function and extracting structure-related features                                      & All                                    & 94.80\%                 \\[0.6cm]
ICA–AAE         & \cite{chen_new_2025}        & Handle nonlinearity, non-Gaussian features & All faults except 3,9,15                                    & 99.52\%                 \\[0.6cm]
DRAE         & \cite{ji2025differential}        & Captures transient, time-varying behaviors. Early-stage fault detection. Also a real-world chemical plant &         faults 1 to 14 except 3 and 9                           & 92.34\%                 \\[0.6cm]
\bottomrule
\end{tabular}
}
\end{table}

\subsection{Generative Models}

Generative models are machine learning algorithms that learn patterns and distributions of the data to generate new data, including models such as Generative Adversarial Networks (GANs), Variational Autoencoders (VAEs), and diffusion models. GANs are deep learning models utilized for dealing with imbalanced data, as they are generative algorithms that can increase the number of samples and improve data-driven algorithm accuracy by providing more data. GANs consist of two networks, a generator model that tries to generate synthetic samples and a discriminator model that discovers if a sample is generated by the generator or not \cite{goodfellow2020generative}.



\par For solving the imbalance fault diagnosis problem, an enhanced auxiliary classifier generative adversarial network (EACGAN) is introduced \cite{peng2020imbalanced}. In this model, the boundary-seeking loss is used for stabilizing the training process, with the normal to-fault ratio of 10:1 and considering all faults in the TEP dataset except 3, 9, and 15, the accuracy of the model is 77.01\%. High-efficiency GAN model (HGAN) \cite{qin2022high} is a model that integrates Wasserstein GAN and Auxiliary Classifier GAN to improve the stability and efficiency of the training process. This model is not just limited to generating data, but can also be used directly for fault classification. For the balanced TEP dataset, considering all faults, the accuracy is 86.6\%, while by the normal to fault samples ratio of 5:1, the accuracy is 82.0\%.  Sparse semi-supervised GAN (SSGAN) \cite{liu2020fault} is a model that uses unlabeled and labeled data for training the model. In this model, the discriminator is transformed into a multi-dimensional classifier. The loss functions for labeled, unlabeled, and generated data types are defined separately, and the total loss is used in optimization. Lomov et al. \cite{lomov2021fault} studied different recurrent and convolutional networks and introduced a novel temporal CNN1D2D architecture. A GAN is used to enrich the training data samples. This method performs well in the detection of 3, 9, and 15 faults, which are very challenging.
Shirshahi et al. \cite{shirshahi2025intelligent}  introduced a novel architecture for detecting similar patterns and preventing alarm floods in industrial systems. First, process variables and alarm sequences are jointly represented and pre-processed; these fused data are then mapped into a latent space by a VAE. A GAN measures similarity between latent patterns, and multi-sensor information fusion aggregates similarity indices to identify the root cause of the faults. By incorporating probability density functions of the extracted features, the method mitigates the influence of random noise. The online implementation supports early fault detection and rapid operator decision-making, demonstrating efficacy on the TEP simulator. 
Other generative models, such as diffusion models, can be used for data generation to enhance FDD in industrial systems, eliminating the need for adversarial training. In \cite{li2025improved}, an improved supervised contrastive learning (ISCL) with a denoising diffusion probabilistic model (DDPM) is used to address the challenge of overlapped normal and abnormal data distributions.

In summary, generative approaches can address imbalanced and scarce fault data challenges by synthesizing realistic samples for minority fault classes. Augmenting training sets via GANs can significantly improve classification balance, especially for rare faults. VAEs and hybrid VAE–GAN models provide structured latent spaces, assisting in both anomaly detection and interpretability. Diffusion models, though less explored, show promise for stable, high-quality data generation without adversarial training challenges. Yet, generative models often require careful tuning to prevent mode collapse or generation of unrealistic samples, and their benefit is less noticeable when the original dataset is already large and balanced.

\subsection{Deep Belief Network (DBN)}

\par Deep Belief Networks are primarily used for unsupervised learning tasks, especially for feature learning. They are made of multilayer Restricted Boltzmann machines (RBMs), which are energy-based probabilistic models. DBN consists of visible and hidden layers; the visible layer represents the input data, while the hidden layers capture abstract and complex features. Each layer is fully connected to the adjacent layers, but there are no connections within a single layer. 


One of the first studies that used DBN for chemical plants FDD is \cite{xie2015hierarchical}, unlabeled data is used to pre-train the DBN, and then labeled data is used for training the DNN using backpropagation. The average classification accuracy is 80.5\%. Using deep learning techniques for layer-by-layer feature extraction may lead to the weakening or vanishing of the fluctuations that reflect the fault information. To address this issue, in \cite{yu2018layer}, the support vector data description (SVDD) and moving average filter technologies are utilized; moreover, in the structure of DBN, the traditional Restricted Boltzmann Machine (RBM) is replaced with the Gaussian Restricted Boltzmann Machine (GRBM). To reduce overfitting and improve the generalization of DBN, dropout is used in DBN \cite{wei2020research}. Dropout randomly disables a certain portion of units in the training process. By using this method, the average F1-score for all faults of TEP is 0.81. By using deep learning for FDD, important information in the raw data might be filtered during feature compression. To solve this problem, an extended deep belief network (EDBN) is proposed in \cite{wang2020novel}, where each extended restricted Boltzmann machine (ERBM) receives raw data and hidden features as inputs during the pre-training phase. The dynamic properties of process data are then taken into account while building a dynamic EDBN-based fault classifier. The average FDR of all TEP faults except 3, 9, 15 using EDBN is 94.31\%, which is slightly better than DBN 93.89\%. The existing noise level in industrial environments affects the performance of deep learning FDD models. To reduce the noise effects, a penalty factor is added to the network structure, avoiding the local optimal situation of a DBN and increasing FDD accuracy, besides an adaptive lifting wavelet (ALW) that reduces the noise level, leading to the ALW-DBN model\cite{yao2021hybrid}. The average FDR of TEP for all faults is 97.33\%. However, the ALW-DBN model's training takes longer because of the iterative penalty optimization algorithm. Additionally, the DBN method has obvious drawbacks when attempting to extract the time information from chemical data in comparison to the DCNN method. Lui et al. \cite{liu2021industrial} introduced an index for measuring fault information of high-dimensional features and finding deep highly-sensitive features (DHSFs) from the extracted features of a DBN. In the mentioned research, the Gauss-Bernoulli Restricted Boltzmann Machine (GRBM) is used to replace the traditional binary Restricted Boltzmann Machine (RBM) to reduce information loss.  Finally, Euclidean distance (ED) is used to calculate the distance between normal and fault samples, and the moving average (MA) to eliminate the effect of noise. The average FDR of the proposed  DHSF-DBN method is 88.67\%. Yang et al. \cite{yang2021pairwise} proposed a pairwise graph regularized deep belief network (PG-DBN) model for FDD. Two different graph constraints are considered on the hidden layer of the RBM, preserving the feature manifold structure in the same class of the data and penalizing the feature manifold structure in different classes of the data, respectively. The average classification accuracy of PG-DBN, G-DBN, CNN, and DBN, which are studied in this paper, is 93.87\%, 91.98\%, 89.18\%, and 83.1\% for all 21 faults of TEP.

DBNs demonstrate competitive performance when combined with enhancements such as graph regularization (PG-DBN) or noise reduction (ALW-DBN). These methods handle nonlinearities well and can be trained with limited labeled data, making them attractive for industrial contexts where annotation is expensive. However, DBNs are computationally heavier than CNN or AE alternatives and often lag in capturing long temporal dependencies, limiting their performance in processes with slow or gradual fault progression. Recent DBN variants integrate temporal features explicitly, but this area remains less developed compared to RNN or CNN hybrids.

\subsection{Transformers and Self-attention Networks}
\par Transformer models were introduced by Vaswani et al. \cite{vaswani2017attention} in 2017 and are the foundation for many state-of-the-art NLP models. They consist of an encoder and a decoder, and the input and output are sequences. The innovation of this model lies in its attention mechanism, which assigns varying weights to different samples in the input sequence. This enables the model to capture important information from the entire sequence, rather than just a fixed window size. 



\par Fault happens when a variable deviates from the normal condition. Rapid changes can easily be captured by ML models; however, slow changes require a long time series, which can be analysed by attention-based models that have a global view of the whole process. Deep learning models like RNNs and CNNs have been widely applied for FDD of chemical processes; however, they are insufficient in extracting features of long-term dependencies. Self-attention mechanisms and transformer models can perform better for this goal. These models are primarily designed for natural language processing (NLP) tasks; however, time series data are very similar to text sequences \cite{wei2022novel}. One of the first studies that used transformers for chemical process fault detection is \cite{zhang2022generalized}, they introduced a general transformer in which self-attention is replaced by graph attention, and it achieved an accuracy of 80.10\% for 21 fault conditions. Wei et al. \cite{wei2022novel} proposed a Target Transformer algorithm that has both self-attention and target-attention mechanisms. The model retains the original transformer's encoder and applies target attention to the decoder to extract fault-related features. The average FDR for 20 faults of TEP is 90.39\%. Zhou et al. \cite{zhou2023exploring} proposed an Industrial Process Optimization ViT (IPO-ViT) model based on Vision Transformer (ViT) for FDD. In this research, a CNN, which can extract local features and information, is combined with multi-head attention, which is famous for its global attention. FDR of the proposed model is increased compared to CNN and LSTM models, especially for faults 3,9, and 15. The average FDR for 20 faults of TEP is 94.08\%.


 \par Zhang et al. \cite{zhang2019recurrent}, applied the self-attention (SA) mechanism to the gated recurrent unit (GRU), and introduced an encoder-decoder framework for FDD, where the encoder extracts the features and the decoder does the fault detection. The model is validated on TEP, and the average recall rate for GRU-SA is 0.9504, which is better than GRU's 0.9214. However, this approach can only be used when enough and rich fault samples are available. Xiong et al. \cite{xiong2023attention} introduced a deep learning fault diagnosis method based on fully convolutional networks, long-short-term memory, and attention mechanism called ALSTM-FCN for FDD. The CNN and LSTM layers are incorporated for their spatial and temporal feature extraction abilities, respectively, and the attention mechanism is included for its ability to adaptively focus on the influence of important features. Dropout is used to prevent overfitting, and batch normalization is used to avoid the negative effects of covariate shift. Results show that the proposed method performs better in comparison to traditional CNN and LSTM models. The average accuracy for all 20 faults of TEP is 88.80\%, and for faults 3 and 9, it is less than 80\%. 
 
 Attention mechanisms and transformers are emerging tools for long-sequence modeling in industrial CM, outperforming traditional RNN/CNN hybrids when capturing slow drifts and long-range correlations. Vision Transformer–CNN hybrids (IPO-ViT) and target-attention transformers achieve improved detection for challenging faults (3, 9, 15) by combining local feature extraction with global attention. However, these models are data-hungry and can overfit when fault samples are scarce. In such cases, simpler attention-augmented RNNs or CNNs often provide a better accuracy–complexity trade-off. Transformer adoption in CM is still in its early stages, and future work will likely focus on reducing computational demands while improving robustness to noise. A summary of these methods is shown in \ref{tab:others}.


\begin{table}[h!]
\centering
\caption{Summary of DBN and transformers and Attention networks. See Table~\ref{tab:TEPFaults} for fault types.}
\label{tab:others}
\renewcommand{\arraystretch}{1.3}
\resizebox{0.9\textwidth}{!}{
\setlength{\tabcolsep}{5pt}
\begin{tabular}{p{0.16\textwidth}p{0.12\textwidth}p{0.4\textwidth}p{0.14\textwidth}p{0.08\textwidth}}
\toprule
\textbf{Model} & \textbf{Reference} & \textbf{Specification} & \textbf{Faults Information} & \textbf{Average Accuracy} \\
\hline

DBN                 & \cite{wang2020novel}           & Extended deep belief network (EDBN) enhances the data quality during feature compression                                                 & All faults except 3,9,15               & 94.31\%                 \\[0.6cm]
DBN                 & \cite{yao2021hybrid}           & Adaptive lifting wavelet-DBN (ALW-DBN) to reduce the noise level                                                                         & All                                    & 97.33\%                 \\[0.6cm]
DBN                 & \cite{yang2021pairwise}        & Pairwise graph regularized deep belief network (PG-DBN) to monitor and control the feature manifold structure of the data                & All                                    & 93.87\%                 \\[0.6cm]
Transformer         & \cite{zhang2022generalized}    & Self attention transformer                                                                                                               & All                                    & 80.10\%                 \\[0.6cm]
Transformer         & \cite{wei2022novel}            & Target transformer using both self-attention and target-attention mechanism                                                              & 20 faults                              & 90.39\%                 \\[0.6cm]
Transformer         & \cite{zhou2023exploring}       & Vision Transformer and CNN increasing the accuracy, especially for faults 3,9, and 15                                                     & 20 faults                              & 94.08\%                 \\[0.6cm]
Attention mechanism & \cite{zhang2019recurrent}      & Combines self-attention (SA) mechanism with the gated recurrent unit (GRU)                                                                 & 20 faults                              & 95.04\%                 \\[0.6cm]
Attention mechanism & \cite{xiong2023attention}      & Attention-based LSTM and Fully Convolutional Network (ALSTM-FCN) with Dropout                                                            & 21 faults                              & 88.80\%                 \\
\bottomrule
\end{tabular}
}
\end{table}

\section{Discussion, Challenges and New Frontiers}

This study provides a summary of the most important deep learning algorithms that have been used for condition monitoring of industrial plants. Deep learning is a powerful tool for feature extraction and handling the large and complex data of chemical plants. LSTM and CNN models can find the hidden representations of the data. In general, recurrent networks such as RNN, LSTM, and GRU perform well on time series, sequential data, and data with time dependencies. Transformers and attention mechanisms are also powerful tools for analyzing sequences by weighting the most important samples. Autoencoders reduce the dimension of the data and extract features. Different architectures and models, such as CNNs, can be used to design the encoder and decoder parts. Autoencoders are unsupervised learning methods, and some of the variations, like denoising Autoencoders, are robust against noise. GANs are designed to generate data and can be used to enrich the dataset and increase the accuracy of the condition monitoring. A summary of the discussed methods and their advantages and limitations is provided in Table \ref{tab:algorithms}. 
The following part of this paper discusses the challenges of data-driven condition monitoring. Some challenges, such as unlabeled, unseen, and imbalanced data, are related to the availability of the data, which can be handled by choosing the proper ML algorithm. Using AI and ML algorithms is consistently challenging due to their black-box nature. The system's decisions are not easily interpretable for humans, posing difficulties in establishing trust and hindering real-world implementation. To address this problem, explainable AI, causal AI, and ensemble learning approaches are introduced as a way to increase the accuracy and trust of the decision-making systems.

\begin{table}
    \centering
    \caption{Comparative analysis of intelligent condition monitoring methods.}
    \label{tab:algorithms}
    \resizebox{0.9\textwidth}{!}{
    \begin{tabularx}{\linewidth}{p{0.2\textwidth}p{0.4\textwidth}p{0.3\textwidth}}
        \toprule
        \textbf{Method} & \textbf{Advantages} & \textbf{Limitations} \\
        \midrule
        Recurrent Networks & Process sequential or temporal data, modeling time dependencies & Prone to gradient vanishing/exploding \\[0.6cm]
        Attention Models & Handling sequential data and weighting important samples in the input sequence. No fixed window size is needed. Can capture slow changes in long time series and extract features of long-term dependencies & Needs rich historical data \\[0.6cm]
        CNN & Powerful feature extraction capability. Has denoising capabilities \cite{lu2017intelligent}. & May need multiple layers and large labeled datasets to perform well \\[0.6cm]
        Autoencoder & Feature extraction and dimension reduction ability. Unsupervised training. Robust against noise & Pretraining is needed \\[0.6cm]
        GAN & Generate data to enhance the data imbalance problem. Can be used for semi-supervised classification \cite{qin2022high} & Stability problem in the training \\[0.6cm]
        DBN & Can be trained unsupervised or supervised. Energy-based probabilistic models & Computationally expensive training \\
        \bottomrule
    \end{tabularx}
}
\end{table}

\subsection{Data Challenges}

\begin{figure}[htbp]
     \centering
    \includegraphics[width=0.9\linewidth]{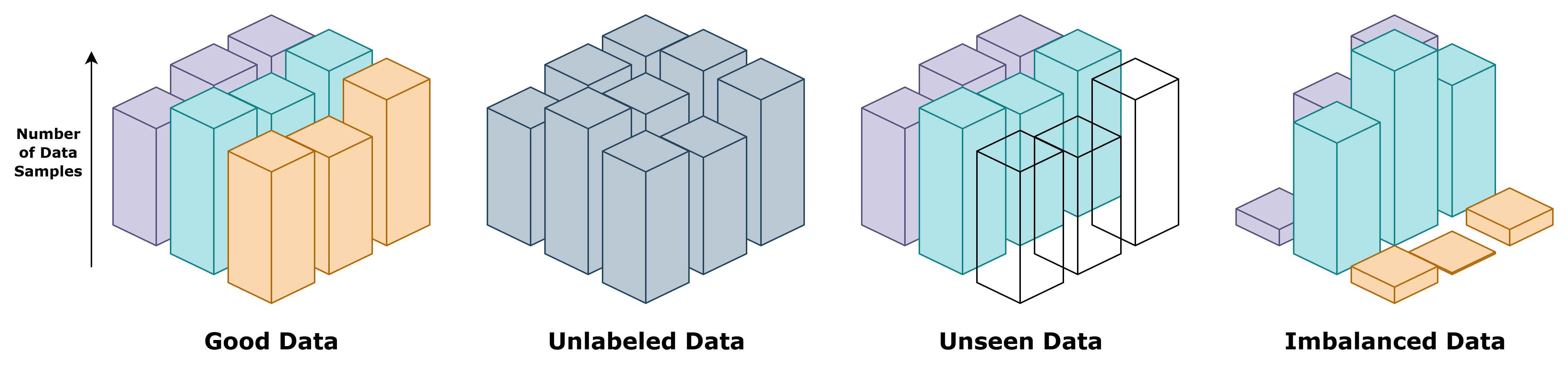}
    \caption{Common data challenges in industrial condition monitoring}
    \label{fig:data-challenges}
\end{figure}

This section addresses common data challenges in industrial environments, namely unlabeled data, unseen fault scenarios, and class imbalance. These issues, illustrated in Figure \ref{fig:data-challenges}, often limit the performance of conventional ML and DL algorithms. To address them, we review advanced strategies specifically designed to mitigate these challenges and thereby improve the robustness and reliability of condition monitoring in industrial applications.


\subsubsection{Unlabeled Data}

\par Collecting and labeling data can be quite challenging, especially in industrial settings. Even when data is properly collected, labeling it may not be straightforward. This labeling process is often time-consuming and expensive, typically requiring the supervision of an expert. Various approaches, such as unsupervised learning and self-supervised learning, have been employed to address this issue. If only the normal data is labeled, anomaly detection methods can be effective.

\par Most FDD methods fail when there is not enough labeled data. Peng et al. \cite{peng2020cost} proposed a method that uses a novel form of the bidirectional gated recurrent unit (BGRU), weighting each training example to overcome the class imbalance and cost-sensitive active learning for selecting the candidate instances to explore the unlabeled data. The proposed method is validated on TEP, with an imbalance ratio of 0.5 and the unlabeled proportion data of 0.6, and achieved an average FDR of 97.5\%. 
Deep learning methods usually need large labeled datasets to achieve good performance. Self-supervised learning is an approach that can learn features from unlabeled data. Pöppelbaum et al. \cite{poppelbaum2022contrastive} proposed a novel self-supervised data augmentation technique based on SimCLR contrastive learning. The goal of contrastive learning is to compare different samples without any labels; the data itself provides supervision. In the SimCLR framework, multiple data augmentation methods for feature representation are used. By only considering 5\% of the available training data, the classification accuracy is 80.80\% for all TEP faults.
To address situations where there are only a few labeled samples of normal data and the remaining available data, including abnormal samples, is unlabeled (positive-unlabeled (PU) learning), a common problem in the industry, a three-step high-fidelity PU (THPU) approach for FDD is introduced in \cite{zheng2022high}. The approach utilizes self-training nearest neighbors (STNN) to generate a positive dataset from unlabeled training data. Subsequently, a positive and negative data recognition (PNDR) algorithm, based on the SAE, augments the positive dataset and generates a negative dataset based on the unlabeled data. Finally, CNN is used as a binary classifier. With only one labeled normal sample, the FDR of the proposed method on TEP is 92.32\% and will increase to 93.75\% with 50 labeled normal samples. Faults 3, 5, 9, 15, and 16, which are very similar to the normal operational state, were not considered in this study. A semi-supervised anomaly detection method composed of an autoencoder and an LSTM model is introduced in \cite{morales2019case}. In this method, a control chart called Open Up is used for fault diagnosis. Open Up provides visual support that enables the interpretation and diagnosis of detected faults in a production system. The average accuracy of the model on a subset of TEP with 6.25\% labeled data is 97.46\%. Gorman et al. \cite{gorman2022anomaly} proposed an interpretable anomaly detection method based on 1-Dimensional Convolutional Autoencoders (1dCAE) and Localised Reconstruction Error (LRE). In LRE, each input channel is evaluated separately, and the sensors and data that lead to anomalies are identified, outperforming global reconstruction errors.
Chadha et al. \cite{chadha2021deep} used a convolutional autoencoder for handling unlabeled data. The latent features are split into discriminative and reconstructive, and an auxiliary loss based on k-means clustering is used for the discriminatory latent variables. The classification average accuracy for 3, 9, 15, and 21 faults, which are very hard to detect, is 50.04\%.
Usually, there are not enough labeled samples, so semi-supervised learning methods are practical as they use unlabeled data to help with the limited labeled data during model training. A consistency regularization autoencoder (CRAE) framework is introduced in \cite{ma2022consistency}, which captures correlations from labeled and unlabeled data. Previous samples are augmented with the current samples to increase the fault detection speed and accuracy. With just 20\% of the dataset's labels, the average fault detection accuracy for all faults in the TE dataset, excluding faults 3, 9, and 15, is 92.95\%.
To utilize unlabeled data more efficiently, Zhang et al. \cite{zhang2022semi}, combined LSTM for temporal feature extraction and ladder autoencoder (LAE) for semi-supervised learning. LSTM-LAE achieves the interpretability to extract fault-relevant process variables. By just having the labels of 11\% of the dataset, this model achieved the average accuracy of 95.9\% for all faults except 3,13, 15, and 16.

Self-supervised contrastive learning extracts information from a large amount of unlabeled data. To enhance the performance of fault diagnosis in scenarios with numerous normal samples and only a few fault classes, a long-tailed distribution is employed in \cite{peng2022progressively}. Progressively balanced supervised contrastive learning (PBS-SCL) is utilized to generate a balanced data batch, facilitating feature learning. Finally, a learnable linear classifier (LLC) is used for classification. The model was validated on a subset of TEP data. With 1.5, 10, 15, 20 samples of faults for each category and sufficient normal samples, the classification accuracy is 47.72\%, 63.92\%, 72.23\%, 86.90\%, and 86.59\%, respectively.

Current zero-shot learning  (ZSL) methods mostly focus on global level fault features; however, local level fault features that characterize the topological relationships between neighboring data points, are very important for generalization, to solve this problem a global–local attentionaware ZSL (GLA-ZSL) method is proposed in \cite{tang_globallocal_2025} consisting of a global prior refinement module using 2D-CNN for global features and local feature enhancement module based on the channel attention and 1D-CN, in addition a meta-learner is used for generalization. Human-defined fault attributes are used in this study.
A dual-network autoencoder-based adversarial domain adaptation with Wasserstein divergence was introduced in \cite{yang2024novel} to diagnose unlabeled data. A dual-network autoencoder combining a convolutional neural network (CNN) and long short-term memory (LSTM) is used to extract both local deep features and temporal information. The autoencoder performs unsupervised reconstruction of source domain data, ensuring high classifier accuracy. An adversarial training process is employed with a domain discriminator to guide the feature extractor in learning domain-invariant features, minimizing the Wasserstein distance to reduce the difference in feature distribution across domains. Additionally, Wasserstein divergence is introduced to the adversarial process to enhance the stability of the training.

\subsubsection{Unseen Data}
 Deep learning-based FDD algorithms can not perform well on unseen operation modes and faults. To deal with previously unknown faults, an algorithm based on supervised contrastive learning (SCL) is used in \cite{peng2022open}, in which the normal samples are contrasted with negative augmentations of themselves. Contrastive Learning is a deep learning technique for unsupervised representation learning. The goal is to learn a representation of data such that similar instances are close together in the representation space, while dissimilar instances are far apart. Soft Brownian Offset (SBO), sampling, and shrinkage autoencoder are used to generate the out-of-distribution (OOD) samples, to obtain some information from unknown classes. The goal is to prevent unseen fault samples from being classified as normal conditions. Balanced fault diagnosis, imbalanced fault diagnosis, and few-shot diagnosis of TEP are tested by the proposed method, demonstrating good performance. Faults 1, 2, 4, 6, and 7 are considered as the seen faults, and faults 8, 12, 14, and 18 are selected as the unseen faults. The detection accuracies in the balanced scenario are 97.5\%, 100\%, 100\%, and 90.63\%. In the imbalanced scenario, where the balance ratio of normal to fault is 10:1, the accuracies are 97.2\%, 98.75\%, 100\%, 90.38\%. In few-shot learning, where there are only 15 samples of each fault, the accuracies are 95.1\%, 97.6\%, 99.86\%, 88.2\%, respectively. Xiao et al. \cite{xiao2022fault} implement a developed domain generalization (DG) method that guarantees generalization over unseen domains. DG learns a universal fault diagnosis model from historical operating modes and can generalize to unseen modes. A novel labeling and class progressive adversarial learning (LCPAL) algorithm for FDD is introduced to learn domain-invariant feature representations. This model aligns the label distribution by weighting the source classification loss based on the number of instances in each category.
 Stacked autoencoders can be used for handling incomplete, missing, or unlabeled data. In \cite{guo2020deep}, a modified stacked autoencoder is used to learn low-feature representations from incomplete data input. Only 40 input variables out of 52 are randomly chosen for fault detection, and the average accuracy for the fault diagnosis of faults 1, 4, 7, 8, 10, 13 is 83.76\%.

 A Conditional Gaussian Network (CGN)-based Bayesian Network (BN) framework is proposed in \cite{atoui2019single} for FDD in industrial processes, with the added capability to handle unknown fault classes through a modified exclusion criterion. Unlike traditional approaches that require separate models for detection and diagnosis, the proposed method employs a single BN to simultaneously perform both tasks. By introducing a probabilistic boundary that respects statistical significance levels, the model reduces false alarms and enables statistical fault classification, including the detection of novel faults not seen during training. The approach is validated using TEP under multiple fault scenarios and compared against benchmark methods. Results show improved classification accuracy and a substantial reduction in false alarm rates. Moreover, the proposed framework is computationally simpler and scalable, making it suitable for complex systems with limited labeled fault data.

\subsubsection{Imbalanced Data}

Industrial systems usually operate under normal condition, hence the normal data is ample. However, faults occur infrequently, and the number of fault samples is rarely equal to the number of normal samples. AI algorithms can not perform well when the data is imbalanced. The performance of the data-driven FDD algorithms depends on the availability of the data; however, in many real-world cases, data is imbalanced or incomplete. To deal with this problem and achieve good FDD accuracy, different approaches are introduced. One approach for handling imbalanced data sets is generating synthetic samples of the minor class to make the data balanced \cite{ahang2022synthesizing}. The other approach involves using feature extraction methods that focus on available data, as well as considering one-class classification and anomaly detection methods. 

\par Tian et al.\cite{tian2020identification}, used a generative adversarial network (GAN) to reconstruct missing data, Spearman's rank correlation coefficient (SRCC) for dimension reduction, select the most important features and noise canceling, and finally DBN for FDD. The average FDR of the process for 20 faults of TEP is 89.70\%, which is 6.23\% higher than the traditional DBN method. Hu et al.\cite{hu2018imbalance} proposed an incremental imbalance modified deep neural network (incremental-IMDNN) in combination with active learning for FDD of imbalanced data. An imbalance modified synthetic minority oversampling technique (IM-SMOTE) is used for addressing data imbalance. The model extends the structure hierarchically with the arrival of new faults. Similar faults are merged into the same classes by using fuzzy clustering. To tackle the class imbalance problem in FDD of chemical processes, an Imbalance Modified Convolutional Neural Network combined with incremental learning, as described in \cite{gu2021imbalance}, generates new samples from imbalanced raw data using a dynamic modified synthetic minority oversampling technique (DM-SMOTE). This model updates itself by receiving the new data. In \cite{amini2022fault}, a source-aware autoencoder (SAAE) for detecting unseen faults and handling imbalanced data sets is introduced that uses Bidirectional long short-term memory (BiLSTM) and residual neural network (ResNet) to deal with randomness. The training can be done using just normal data, and then fine-tuned by using emerging faulty samples. The proposed method detects unseen faults with an accuracy of 93.8\%; moreover, the average accuracy of fault detection for different fault-to-normal ratios is 91.8\%. In the balanced condition, the average accuracy is 94.3\%. 
Peng et al. \cite{peng2022non} introduced a non-revisiting genetic cost-sensitive sparse autoencoder(NrGCS-SAE) for solving the imbalanced fault diagnosis problem. The genetic algorithm optimizes class weights while the non-revisiting strategy increases exploration ability. This method could detect faults in the unbalanced TE dataset with an average accuracy of 92.02\% while the normal-to-fault ratio was 10:1.


To identify unknown faults, a few-shot learning-based unknown recognition and classification (FSLB-UR\&C) approach is introduced in \cite{mirzaei2023identification}. For feature extraction and comparing the similarity of features, a gated recurrent unit with a dot-product attention layer is used. This method does not require a large amount of data for training. The classification accuracy of the known faults of TEP is 88.9\%, and for unknown faults is 85.7\%. The known faults include Faults 0, 2, 12, and 14, while the unknown faults include Faults 1, 4, 5, 7, 8, 10, 11, and 13. Xu et al. \cite{xu2023new}, propose a few-shot learning fault diagnosis method based on a class-rebalance strategy, to diagnose new scarce faults. The fault samples are transferred into a feature space, where each type of fault, even each new instance, is mapped into a separate feature cluster. In this way, fault diagnosis can be achieved by estimating feature similarity between instances and faults. 1D-CNN with an attention module is used for feature extraction. Data augmentation and class rebalance are done to increase the fault samples of scarce classes. Considering 3 to 50 samples of faults for each class the average diagnosis accuracy of the proposed model on TEP is from 81.8\% to 94.7\%.

Zhang et al. \cite{zhang2023effective}proposed a ZSL approach to handle unseen faults. Zero-shot learning involves the utilization of expert-provided high-level descriptions instead of relying on trained objects to identify target items. These descriptions encompass semantic attributes such as colour, shape, and behaviors, which can be learned beforehand without requiring samples from unfamiliar categories. In essence, ZSL is an approach to train models to recognize patterns in new types of data using information gleaned from familiar classes along with relevant descriptions. Initially, feature extraction from raw data is performed using a one-dimensional convolutional neural network (1D CNN), followed by labeling with semantic descriptions based on expert knowledge. Five Different subsets of TEP are used to validate the proposed method, considering different faults as unseen data. The detection accuracies for different scenarios vary from 59.72\% to 96.67\%, which outperforms other ZSL methods like SAE in most of the scenarios. 


\par To address the class imbalance problem, the double branch rebalanced network (DBRN) method is proposed in \cite{ma2023double}. At first, a resampling method rebalances the classes, then a cost adaptive reweighting strategy rebalances the cost of each branch. Finally, the two aforementioned steps are merged by a fusing learning strategy. The imbalance ratio is 10:1, and for faults 1,2,6 of TEP the average diagnosis accuracy is 98.75\%.

\par
\begin{figure}
\centering
\includegraphics[scale=0.9]{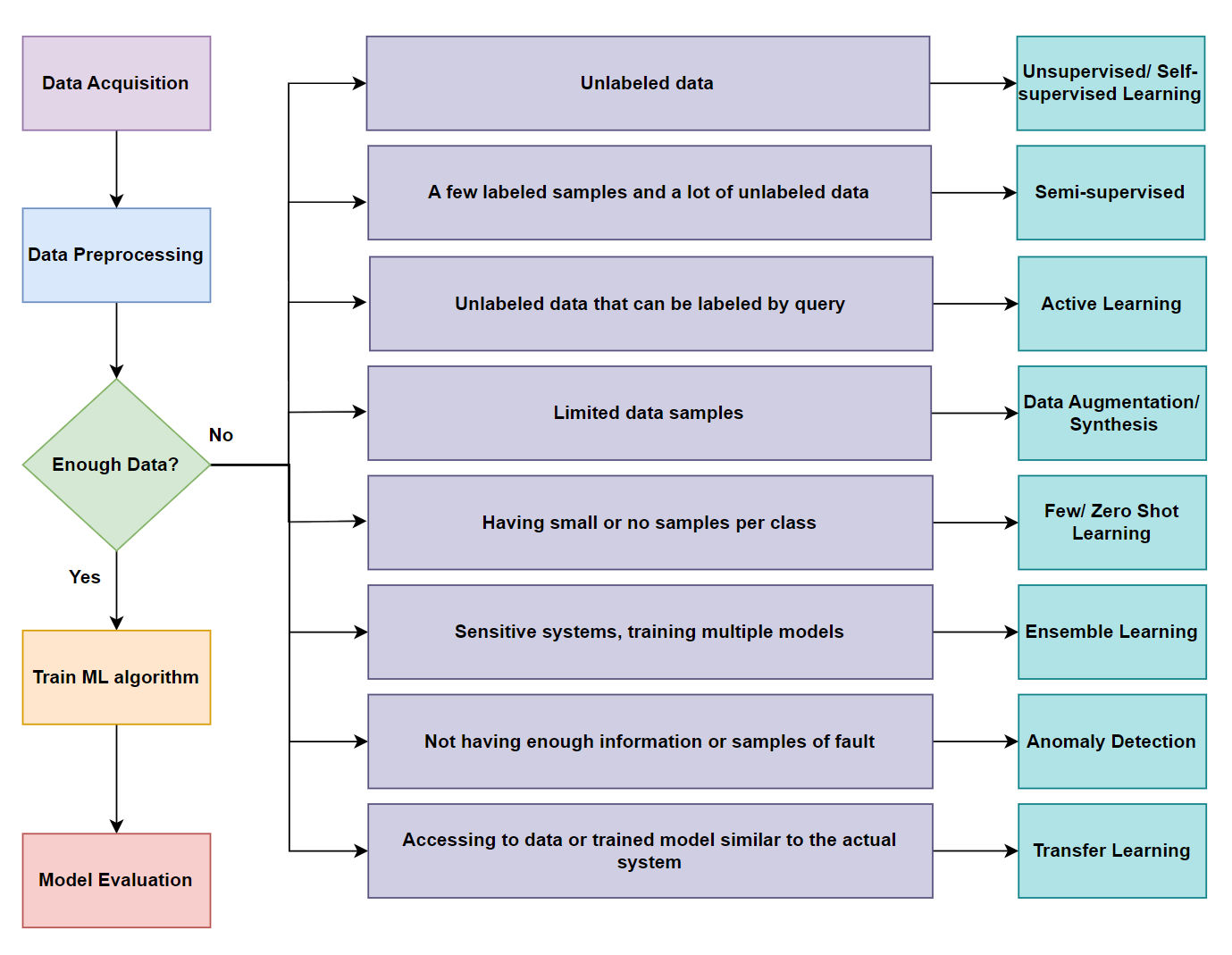}
\caption{Process of data-driven fault detection considering data limitation \label{FDD}}
\end{figure}

 Guo et al. \cite{guo2020deep}, used a modified stacked autoencoder to learn low-level features containing the information of complete data based on incomplete input data. This method creates data representations that can be used for classification.
The strategies for the data-driven fault detection process are shown in \ref{FDD}. When the data is not rich, two general approaches can handle the problem. Augmenting or generating more data, or using the models that can perform with few samples. Different scenarios and possible solutions are shown in the mentioned figure. Table \ref{tab:Imb} shows the summary of methods that can handle imbalanced, incomplete, or unlabeled data.

\begin{table}[h!]
\centering
\caption{Solutions to imbalanced and unlabeled data}
\label{tab:Imb}
\renewcommand{\arraystretch}{1.25}
\resizebox{0.9\textwidth}{!}{
\setlength{\tabcolsep}{4pt}
\begin{tabular}{p{0.17\textwidth}p{0.12\textwidth}p{0.40\textwidth}p{0.17\textwidth}p{0.09\textwidth}}
\toprule
\textbf{Method} & \textbf{Reference} & \textbf{Specification} & \textbf{Fault Information} & \textbf{Avg. Accuracy} \\
\midrule
unlabeled &\cite{morales2019case} & Combination of autoencoder and LSTM model and a control chart for anomaly detection of unlabeled data & 6.25\% of the data is labeled & 97.46\% \\ [1cm]
unlabeled & \cite{peng2020cost} & Bidirectional gated recurrent unit (BGRU), weighting training example to overcome the class imbalance and cost-sensitive active learning to explore the unlabeled data & Data imbalance ratio of 0.5 and the unlabeled proportion data of 0.6 & 97.50\% \\ [1.5cm]
unlabeled and Imbalanced &\cite{peng2020imbalanced} & Enhanced auxiliary classifier generative adversarial network (EACGAN) to deal with imbalanced data and more stable training & Normal to fault ratio of 10:1, all faults except 3, 9 and 15 & 77.01\% \\[1cm]
Imbalanced & \cite{zheng2022high} & Three-step high-fidelity positive-unlabeled (THPU) is used to deal with having a few labels in the data set, by generating and augmenting data & One labeled normal sample is available the rest is unlabeled & 92.32\% \\[1.5cm]
unlabeled and Imbalanced & \cite{peng2022progressively} & Contrastive learning for information extraction of unlabeled and imbalanced data & Only 20 samples of each fault is available & 86.59\% \\
Imbalanced &\cite{amini2022fault} & Source-aware autoencoder (SAAE) for handling imbalanced data combining BiLSTM and ResNet & Detecting unseen faults & 93.80\% \\
unlabeled &\cite{ma2022consistency} & Consistency regularization autoencoder (CRAE) for handling unlabeled data & All faults except 3, 9, and 15 (20\% of data is labeled) & 92.95\% \\
unlabeled and Imbalanced & \cite{peng2022non} & Using genetic algorithm and sparse autoencoder for imbalanced and unlabeled fault diagnosis & unlabeled data and the normal to fault ratio of 10:1 & 92.02\% \\ [1cm]
Imbalanced &\cite{qin2022high} & High-efficiency GAN using Wasserstein GAN and Auxiliary Classifier, to generate and classify data for an imbalanced dataset & Normal to fault ratio of 5:1 & 82.00\% \\[1cm]
unlabeled &\cite{zhang2022semi} & Combined LSTM for temporal feature extraction and ladder autoencoder (LAE) to handle unlabeled data & All faults except 3,13, 15, and 16, 11\% of the dataset is labeled & 95.90\% \\[1cm]
Unknown Faults & \cite{mirzaei2023identification} & Few-shot learning and GRU with attention layer for unknown fault identification & Faults 0, 2, 12, and 14 as known faults, Faults 1, 4, 5, 7, 8, 10, 11, and 13 as unknown & 85.70\% \\[1cm]
Imbalanced &\cite{xu2023new} & Few-shot learning based on a class-rebalance strategy to detect new faults. Feature extraction is done using 1D-CNN with an attention module & 3 to 50 samples of faults for each class & From 81.8\% to 94.7\% \\[1.5cm]
Imbalanced & \cite{ma2023double} & Double branch rebalanced network (DBRN) to deal with imbalanced classes using resampling and reweighting & Imbalance ratio is 10:1 for faults 1,2,6 & 98.75\% \\

\bottomrule
\end{tabular}
}
\end{table}

\subsection{Emerging Methods and Future Directions}

This section highlights emerging methodologies and concepts that have significant potential for improving intelligent condition monitoring. While these approaches have demonstrated promising results, they remain relatively underexplored in industrial applications. By examining these cutting-edge approaches, we aim to identify future research opportunities that can address current limitations, leading to the development of more reliable, interpretable, and scalable condition monitoring systems.

\subsubsection{Ensemble Learning}

Ensemble learning is a technique that combines the predictions of multiple models to improve the overall performance and robustness of a predictive model. The idea behind ensemble learning is that by aggregating the outputs of several models, you can often achieve better results than any single model can produce \cite{sagi2018ensemble}. Ensemble methods work by creating a committee of base models, which can be of the same or different types, and then combining their outputs in various ways. Ensemble learning improves the accuracy, robustness, and stability of the model.
Plakias et al. \cite{plakias2022novel} introduced an unsupervised fault detection approach by the ensemble of different autoencoders, each trained for detecting one type of fault. A soft voting process makes the final decision. In this approach, an anomaly indicator is generated for unknown samples, showing the confidence of the network regarding the condition of the system.
To improve the feature learning for complex, nonlinear, and high-dimensional signals, a hybrid model of integration of 1-DCNN and stacked denoising auto-encoders (SDAE) is introduced in \cite{zhang2021fault}. This method has higher classification accuracy compared to 1-DCNN, SDAE, and DBN.

Li et al. \cite{li2022ensemble} proposed a novel ensemble monitoring framework for industrial process data to address the complexities inherent in industrial processes, which exhibit various characteristics such as linearity, non-linearity, Gaussian, non-Gaussian, and dynamic behavior. The framework aims to automatically determine the most suitable local models and optimal monitoring variables by using a combination of multiple statistical analysis methods. The key contributions of the study include the automatic determination of local models and monitoring variables, the use of multiple models to capture diverse data characteristics, and the elimination of redundant models to enhance monitoring performance. An ensemble convolutional neural network is introduced in  \cite{najaran2023evolutionary}, which combines base learner algorithms trained on features extracted by different approaches such as Fourier transforms, Wavelet transforms, Walsh transforms, and Hilbert–Huang transforms. CNNs can also automatically extract features from these signals. The output of base learners is used in a weighted voting paradigm, which is optimized via an evolutionary algorithm.
Typically, multiple ML algorithms are combined in ensemble learning; However, as a future direction of research for complex systems like TEP, ensemble learning could be used to aggregate model-based and signal-based methods. This approach could increase the robustness of the condition monitoring systems against uncertainties like imbalanced, incomplete, or unlabeled data.

 \subsubsection{Hybrid and Informed Machine Learning}

Data-driven models highly depend on the availability and quality of the data, which is very difficult to acquire in complex industrial setups such as chemical plants. By integrating prior knowledge into data-driven and ML methods, and utilizing hybrid and informed approaches, we can overcome the data scarcity and enhance the accuracy and safety of the process \cite{wu2024physics, wilhelm2021overview}. Hybrid FDD approaches that combine model-based reasoning with data-driven learning are gaining attention in the context of safety-critical industrial systems. These methods leverage both the interpretability of physical or probabilistic models and the adaptability of machine learning techniques. Informed machine learning describes learning
from a hybrid information source that consists of data and prior knowledge. The prior knowledge can be integrated into machine learning in several stages, including training data, hypothesis set, learning algorithm, or final decision \cite{von2021informed}.

Alauddin et al. \cite{alauddin2023integrating} integrate process dynamics into data-driven models to improve the model generalization when the available data is sparse. An additional layer is added to the neural network architecture to incorporate equations and field knowledge, introducing a process dynamics-guided neural network (PDNN) model. Compared to purely data-driven neural networks, PDNN shows improvement on reduced sample-sized data. However, an accurate model of the process dynamics is needed to achieve acceptable performance, and such models are computationally more expensive.  
 Chao et al. \cite{chao2022fusing} introduce a hybrid framework using physics-based models to infer unobservable model parameters, which are later combined with sensor readings and used as an input to a deep neural network for prognostics of complex systems. The framework was implemented for turbofan engine remaining useful lifetime (RUL) prediction, and outperformed purely data-driven approaches and required less training data. Tidriri et al. \cite{tidriri2018generic} introduced a framework for decision fusion of heterogeneous methods for FDD to obtain reliable results. The proposed discrete Bayesian Network (BN) approach is validated using TEP, combining model-based and data-driven approaches, and outperforms the individual decisions, achieving an FDR of 99.21\%.
 

\subsubsection{Explainability}

AI models have become crucial tools for condition monitoring of industrial processes, providing unprecedented advancements in efficiency optimization, prediction performance, and data-driven decision making \cite{yan2023review}. However, the integration of AI has not been without drawbacks. A significant concern arises from the inherent complexity of AI models, often treated as black boxes due to their inexplicable internal mechanisms. The inability to decipher the rationale behind AI predictions and decisions has led to a lack of trust, hindering their effective real-world implementation. In industrial contexts, where decisions have tangible consequences, the absence of interpretability becomes a critical barrier \cite{zio2022prognostics}. Explainable AI (XAI) methods address the interpretability issues of AI and other black-box models by providing end-users with transparent and understandable information regarding the model's internal decision mechanisms. 

XAI is an area of research focused on clarifying the intricate functionality of AI models and illuminating their decision-making processes. In essence, XAI bridges the gap between complex AI methodologies and human comprehensibility. By examining a model's outputs and unraveling its internal workings, XAI techniques empower users with insights into the rationale behind AI-generated outcomes. Many XAI techniques have been explored in the literature. However, they can generally be classified as either intrinsic approaches, where the model is designed or refined to be particularly interpretable by humans, or post-hoc approaches, where external tools are applied in parallel to analyze the model's internal mechanisms with respect to predicted outputs \cite{nor2021overview}. Intrinsic methods, sometimes referred to as interpretable AI, are concerned with the development of the AI model itself, limiting the design to simple methods such as linear regression and logic-based rules. Post-hoc methods, on the other hand, can be model-specific or general to any Black box model with varying levels of explanations, either locally or globally \cite{confalonieri2021historical}. Commonly used post-hoc XAI tools include local interpretable model-agnostic explanations (LIME) \cite{ribeiro2016should}, which works by learning local interpretable models around the prediction, and Shapley additive explanations (SHAP) \cite{lundberg2017unified}, which uses ideas from game theory to explain model contributions.

In the study by Bhakte et al. \cite{bhakte2022explainable}, an XAI-based condition monitoring framework is introduced specifically for understanding the predictions made by deep learning-based FDD models. The application of this research provides two main benefits: it enhances user confidence by providing explanations for correct fault identifications, and it empowers model developers with insights when the model identifies faults incorrectly. Their methodology, based on the Shapley value framework, uses integrated gradients to identify the variables influencing DNN predictions, which are often related to the root causes of faults. A similar study \cite{peng2022towards} presents a comprehensive two-step framework for the detection and diagnosis of faults in industrial processes, with an emphasis on providing better understandability. The first step leverages a denoising sparse autoencoder to acquire a robust and adaptable feature embedding, while the subsequent phase employs smooth integrated gradients to unveil the significance of features tied to detected faults.
Additionally, \cite{agarwal2021explainability} presents an innovative deep learning-based FDD method that harnesses XAI to refine input features based on their relevance. By utilizing the layerwise relevance propagation (LRP) algorithm, this approach not only eliminates irrelevant input variables but also offers potential insights into the root causes of detected faults. Rule-based classification methods like Logical Analysis of Data (LAD) are used for interpretable fault detection in complex industrial processes like TEP \cite{ragab2018fault}, where hidden knowledge is discovered by revealing interpretable patterns linked to underlying physical phenomena. The decision model explains the potential causes of these faults. Another method to increase the interpretability and explainability of CM, especially for complex systems, is using physics-informed machine learning (PIML) approaches \cite{wu2024physics} that combine physics-based modeling and data-driven machine learning. Collectively, these studies underscore the profound impact of XAI in enhancing the transparency, dependability, and interpretability of condition monitoring models, ultimately yielding more effective fault detection, diagnosis, and comprehension.

To enhance the explainability of Graph Convolutional Network (GCN)-based fault diagnosis methods in industrial systems, Graph-SRGA, a two-stage variable contribution analysis framework, is proposed in \cite{xu2025novel}. Traditional GCN-based approaches treat fault diagnosis as a pure classification task, often neglecting the interpretability of which variables or sensor nodes most influence diagnostic outcomes. This restricts industrial deployment due to reliability concerns.
Graph-SRGA introduces a two-stage analysis process: Neighbor-stage contribution analysis assesses the impact of neighboring nodes in the graph. Sensor-stage contribution analysis evaluates the role of individual sensors connected to these nodes.
The method employs a self-reproductive non-dominated sorting genetic algorithm (SRGA), guided by two objective functions, to identify influential variable subsets that contribute most significantly to the GCN's decision-making. Experimental validation on the TEP and high-speed train traction systems confirms that Graph-SRGA provides accurate, interpretable insights into the contributions of various features, offering a practical solution for improving the transparency and trustworthiness of GCN-based diagnostic models in industrial applications.

\subsubsection{Causality}

Causality plays a key role in the condition monitoring of industrial processes, offering a better understanding of the true cause-and-effect relationships within complex systems. While XAI provides insights into the reasoning and inherent behavior of intelligent models and their outputs, it neglects to consider the true physical systems that generate the data. In contrast, causality is an area of research that specifically focuses on understanding the true causal mechanisms driving physical systems rather than simply relying on statistical associations between variables \cite{pearl2009causality}. These underlying relationships hold particular significance for industrial processes. Industrial processes are often characterized by a multitude of process variables and parameters, coupled with external disturbances and noise. Consequently, data-driven methods are highly susceptible to spurious correlations and erroneous associations. Thus, the exploration of causal analysis becomes critical for analyzing intricate industrial processes.

Causal modeling is the formal representation of these cause-and-effect relationships, often expressed through Bayesian networks or structural equation models. These models are generally crafted by domain experts using their expertise and leveraging their intimate knowledge of the systems. However, when domain knowledge is unavailable, causal models can be approximated via randomized control trials and experimental interventions. Even so, analyzing industrial processes in these ways might prove unfeasible or unsafe. To circumvent this problem, causal discovery methods, such as those examined by Vukovic et al. \cite{vukovic2022causal}, become indispensable. These methods employ statistical algorithms to analyze historical observational data and infer causal relationships among the complex web of variables.

The integration of causality into data-driven AI referred to as Causal AI (CAI), can provide more robust and interpretable model predictions. Since causality represents the fundamental data generation mechanisms of a system, CAI models are less biased toward training data and are more effective in new situations by applying causal generalizations from past observations \cite{cox2023causally}. Causality-informed AI methods can provide more actionable insights by pinpointing the actual causes of anomalies or issues, facilitating targeted and effective responses. Additionally, CAI approaches can reduce false positives/negatives by distinguishing between spurious correlations and true causal links. Furthermore, these methods can be easily adapted for root-cause analysis, an important aspect of fault diagnosis and condition monitoring.

In the context of industrial processes, where the consequences of incorrect diagnoses can be significant, causality becomes a cornerstone. CAI enables accurate identification of root causes and provides a clear understanding of the propagation of faults, making it an indispensable tool for robust condition monitoring.

Several notable studies demonstrate the practical application of CAI in industrial process condition monitoring. For instance, Ma et al. \cite{ma2020novel} have developed a KPI-oriented hierarchical monitoring framework that utilizes Bayesian fusion and a Granger causality method to enhance fault detection and illuminate propagation paths. Similarly, Wu et al. \cite{wu2023three} propose a multimodal monitoring framework for fault detection and propagation analysis using a deep local adaptive network (DLAN) alongside a dedicated causal Bayesian network. Li et al. \cite{li2021fault} employ a spatial sequence-based deep learning approach, leveraging LSTM models with attention mechanisms to predict faults and identify causal variables with remarkable accuracy. Arunthavanathan et al. \cite{arunthavanathan2022autonomous} present an autonomous framework for online monitoring that continuously updates itself, integrating causality through variable contributions extracted by a permutation algorithm. Meanwhile, Bi et al. \cite{bi2023large} introduce a novel causal discovery method centred on the causality-gated time-series transformer (CGTST) model, explicitly identifying causal variables and refining relationships through permutation feature importance (PFI). In contrast, Wang et al. \cite{wang2023root} harness causality with their temporal registration network (TRN), demonstrating improved root cause localization and propagation path identification. These select studies demonstrate the potential of integrating causality to advance the field of industrial process condition monitoring, offering comprehensive insights for more effective fault detection and diagnosis. 

\subsection{Industrial Applicability Beyond TEP}

While the TEP benchmark provides a standardized evaluation, its synthetic nature cannot fully replicate noise patterns, unmodeled dynamics, and operational variability of actual plants. However, data-driven methods have demonstrated strong generalizability, with models developed for one case study often transferable to others with minimal adaptation while maintaining high performance. For example, MRJDCNN-LSTM, introduced in \cite{chen_reinforced_2024}, demonstrated a high accuracy of 97.47\% when tested on the F101/3 coking furnace. Similarly, the GLACNN \cite{yang2025fault} validated on penicillin fermentation processes, achieving an FDR of 97.5\%. The 1-DCNN + SDAE introduced in \cite{zhang2021fault} was validated by the fed-batch fermentation penicillin process (FBFP), and a real-life manufacturing case of an industrial conveyor belt achieving accuracies of 99.68\% and 95.0\%, respectively. The effectiveness of the unsupervised data mining method introduced in \cite{zheng2020new} was also evaluated by an industrial hydrocracking process. The unsupervised ensemble of autoencoders \cite{plakias2022novel} was tested on the Case Western Reserve University (CWRU), the Intelligent Maintenance
 System (IMS) bearing datasets alongside the TEP. BRNN \cite{sun2020fault} was validated using a real chemical manufacturing dataset, specifically an amine tower, demonstrating its superiority in extracting data patterns from complex real-world chemical plants.

\section{Conclusions}

This paper provides a comprehensive review of the application of machine learning (ML) and deep learning (DL) techniques for condition monitoring in industrial plants, with a particular emphasis on chemical processes and the Tennessee Eastman Process (TEP) benchmark. The survey highlights the growing dominance of DL-based methods, which have demonstrated superior capability in handling complex, high-dimensional process data and in automating feature extraction—reducing reliance on manual engineering and domain-specific expertise.
Key challenges frequently encountered in real-world industrial monitoring, including imbalanced and unlabeled data, are critically discussed alongside strategies such as advanced data augmentation, semi-supervised learning, and ensemble methods. The study also highlights the importance of explainability and causal reasoning in AI-driven monitoring systems, as these elements are essential for enabling trust, interpretability, and adoption in safety-critical domains.
From a practical standpoint, this work emphasizes that the effective deployment of AI-based condition monitoring has direct implications for process safety, risk management, and operational efficiency. By outlining emerging solutions and research directions, this review not only consolidates current knowledge but also identifies pathways toward robust, transparent, and industrially viable monitoring frameworks.
In conclusion, intelligent condition monitoring supported by DL and ML represents a transformative approach for industrial systems, offering improved fault detection, reduced downtime, enhanced safety, and greater reliability. Continued efforts in addressing uncertainty, improving explainability, and validating hybrid and transferable models across diverse plants will be pivotal for advancing the next generation of industrial monitoring solutions.

\section*{Declaration of interest}
The authors declare that they have no competing financial interests that could have influenced the work reported in this paper.
\section*{Acknowledgement}
We would like to acknowledge the financial support of FortisBC. and Natural Sciences and Engineering Research Council (NSERC) Canada under the Alliance Grant ALLRP 557088 – 20 in this research.

\printcredits

\newpage

\bibliographystyle{elsarticle-harv} 

%

\bibliography{cas-refs}

\end{document}